\documentclass{article}

\PassOptionsToPackage{numbers,square,sort&compress}{natbib}
 
\usepackage[final]{neurips_2020}

\usepackage[utf8]{inputenc}
\usepackage[T1]{fontenc}
\usepackage[colorlinks=true,citecolor=black,urlcolor=black,linkcolor=black]{hyperref}
\usepackage{url}
\usepackage{booktabs}
\usepackage{amsfonts}
\usepackage{nicefrac}
\usepackage{microtype}
\usepackage[normalem]{ulem}
\usepackage{amsmath,amssymb}

\usepackage{tabularx}
\usepackage{array}
\usepackage{colortbl}
\newcommand{\PreserveBackslash}[1]{\let\temp=\\#1\let\\=\temp}
\newcolumntype{C}[1]{>{\PreserveBackslash\centering}p{#1}}
\newlength{\tblw}

  \newcommand{\T}{^\mathsf{T}}

  \providecommand{\norm}[1]{\|#1\|}
  \providecommand{\innerp}[1]{\langle#1\rangle}

  \providecommand{\op}[1]{\mathcal{#1}}

  \newcommand{\GP}{\mathcal{GP}}
  \newcommand{\dd}{\,\mathrm{d}}

  \usepackage{bm}
  \newcommand{\mathbold}[1]{\bm{#1}}
  \newcommand{\mbf}[1]{\mathbf{#1}}
  \newcommand{\vect}[1]{\mbf{#1}}
  \newcommand{\vectb}[1]{\bm{#1}}

\newcommand{\vomega}[0]{\mathbold{\omega}}

\renewcommand{\mid}[0]{\,|\,}

\newcommand{\Laplace}[0]{\mathcal{L}}
\newcommand{\imag}[0]{\mathrm{i}}

\newcommand{\vf}{\mbf{f}}

\newcommand{\vh}{\mbf{h}}

\newcommand{\vr}{\mbf{r}}

\newcommand{\vw}{\mbf{w}}
\newcommand{\vx}{\mbf{x}}
\newcommand{\vy}{\mbf{y}}

\newcommand{\MK}{\mbf{K}}

  \newcommand{\eg}{\textit{e.g.}}
  
  \newcommand{\cf}{\textit{cf.}}
  \newcommand{\etc}{\textit{etc.}}
  \newcommand{\etal}{\textit{et~al.}}

  \usepackage{booktabs}

  \usepackage{tikz,pgfplots}
  \usetikzlibrary{plotmarks,shapes,arrows}
  \pgfplotsset{compat=newest} 
  \pgfkeys{/pgf/number format/.cd,1000 sep={}}

  \newlength\figureheight
  \newlength\figurewidth

  \usetikzlibrary{external}

  \usepackage[capitalize,nameinlink]{cleveref}
  \crefname{section}{Sec.}{Sects.}
  \crefname{proposition}{Prop.}{Props.}
  \crefname{lemma}{Lem.}{Lems.}
  \crefname{model}{Mod.}{Mods.}
  \crefname{appendix}{App.}{Apps.}

  \makeatletter
  \let\MYcaption\@makecaption
  \makeatother

  \usepackage[font=footnotesize]{subcaption}

  \makeatletter
  \let\@makecaption\MYcaption
  \makeatother
  
  \renewcommand{\paragraph}[1]{{\bf #1}~~}

\definecolor{mycolor0}{rgb}{0.2667,0.4471,0.7098}
\definecolor{mycolor1}{rgb}{0.1647,0.6706,0.3804}
\definecolor{mycolor2}{rgb}{0.8275,0.2627,0.3059}
\definecolor{mycolor3}{rgb}{0.5216,0.4392,0.7176}
\definecolor{mycolor4}{rgb}{0.8118,0.7255,0.4118}
\definecolor{mycolor5}{rgb}{0.2745,0.7176,0.8157}

\definecolor{mylcolor0}{rgb}{0.6902,0.7686,0.8863}
\definecolor{mylcolor1}{rgb}{0.5451,0.8902,0.6941}
\definecolor{mylcolor2}{rgb}{0.9412,0.7490,0.7647}
\definecolor{mylcolor3}{rgb}{0.8627,0.8392,0.9176}
\definecolor{mylcolor4}{rgb}{0.9569,0.9373,0.8667}
\definecolor{mylcolor5}{rgb}{0.7529,0.9020,0.9373}
\definecolor{mylcolor6}{rgb}{0.8750,0.8750,0.8750}

\pgfplotsset{every axis/.append style={grid style={line width=0.6pt,dotted,gray}}}

\title{Stationary Activations for Uncertainty \\ Calibration in Deep Learning}

\author{
  Lassi Meronen \\
  \small Aalto University\,/\,Saab Finland~Oy\\
  \small Espoo, Finland\\
  \texttt{\small lassi.meronen@aalto.fi}

  \And
  \hspace*{-1em}Christabella Irwanto \\
  \hspace*{-1em}\small Aalto University \\
  \hspace*{-1em}\small Espoo, Finland \\
  \hspace*{-1em}\texttt{\small christabella.irwanto@aalto.fi}

  \And
  Arno Solin \\
  \small ~~~~~~~~Aalto~University~~~~~~~~ \\
  \small Espoo, Finland \\
  \texttt{\small arno.solin@aalto.fi}

}

\begin{document}

\maketitle

\begin{abstract}
  We introduce a new family of non-linear neural network activation functions that mimic the properties induced by the widely-used Mat\'ern family of kernels in Gaussian process (GP) models. This class spans a range of locally stationary models of various degrees of mean-square differentiability.  We show an explicit link to the corresponding GP models in the case that the network consists of one infinitely wide hidden layer. In the limit of infinite smoothness the Mat\'ern family results in the RBF kernel, and in this case we recover RBF activations. Mat\'ern activation functions result in similar appealing properties to their counterparts in GP models, and we demonstrate that the local stationarity property together with limited mean-square differentiability shows both good performance and uncertainty calibration in Bayesian deep learning tasks. In particular, local stationarity helps calibrate out-of-distribution (OOD) uncertainty. We demonstrate these properties on classification and regression benchmarks and a radar emitter classification task.
\end{abstract}

\section{Introduction}
Deep feedforward neural networks (see, \eg, \cite{Haykin:1999,goodfellow2016deep}) have become an essential component of modern machine learning. Their black box nature results in a lack of interpretability, an issue that has been tackled recently from many directions, one of which is the study of random (untrained) networks in order to examine what prior assumptions they impose over functions. By assuming a probability distribution on the network parameters, a distribution is induced from the inputs to the outputs of the network. While we typically want networks to have high modelling capability (or \emph{flexibility}), large networks can be hard to analyse directly. This difficulty motivates instead the study of the limiting behaviour of the networks, which can provide new insight and better interpretation. This is also of interest for {\em Bayesian deep learning}, where the concept of assigning priors on neural networks only makes sense if the effects of prior assumptions can be understood.

\begin{figure}[!t]
  \scriptsize
  \pgfplotsset{hide axis,scale only axis,width=\figurewidth,height=\figureheight}
  \setlength{\figurewidth}{.19\textwidth}
  \setlength{\figureheight}{\figurewidth}  
  \begin{subfigure}[t]{.195\textwidth}
    \centering
    \input{./fig/GPbaseline_step.tex}\\[0em]
    ArcCos-0 kernel \cite{cho+saul:2009}

  \end{subfigure}
  \hfill
  \begin{subfigure}[t]{.195\textwidth}
    \centering
    \input{./fig/GPbaseline_relu.tex}\\[0em]
    ArcCos-1 kernel\cite{cho+saul:2009}

  \end{subfigure}
  \hfill
  \begin{subfigure}[t]{.195\textwidth}
    \centering
    \input{./fig/GPbaseline_erf.tex}\\[0em]
    ERF-NN kernel \cite{williams97computing}

  \end{subfigure}
  \hfill
  \begin{subfigure}[t]{.195\textwidth}
    \centering
    \input{./fig/GPbaseline_gauss.tex}\\[0em]
    RBF-NN kernel \cite{williams97computing}

  \end{subfigure}
  \hfill
  \begin{subfigure}[t]{.195\textwidth}
    \centering
    \input{./fig/GPbaseline_matern52.tex}\\[0em]
    Mat\'ern-$\frac{5}{2}$ kernel

  \end{subfigure}  
  \\
  \begin{subfigure}[t]{.195\textwidth}
    \centering
    \input{./fig/MLP_MCdrop_step.tex}\\[0em]
    MLP with step activation

  \end{subfigure}
  \hfill
  \begin{subfigure}[t]{.195\textwidth}
    \centering
    \input{./fig/MLP_MCdrop_relu.tex}\\[0em]
    ReLU activation

  \end{subfigure}
  \hfill
  \begin{subfigure}[t]{.195\textwidth}
    \centering
    \input{./fig/MLP_MCdrop_erf.tex}\\[0em]
    ERF (sigmoidal) activation

  \end{subfigure}
  \hfill
  \begin{subfigure}[t]{.195\textwidth}
    \centering
    \input{./fig/MLP_MCdrop_gauss.tex}\\[0em]
    RBF activation

  \end{subfigure}
  \hfill
  \begin{subfigure}[t]{.195\textwidth}
    \centering
    \input{./fig/MLP_MCdrop_matern52.tex}\\[0em]
    Mat\'ern-$\frac{5}{2}$ activation \\ (this paper)
  \end{subfigure}\\[-10pt]
  \caption{Illustrative comparisons on the {\em Banana} classification data set. The top row shows decision boundaries and marginal predictive variance (low~\protect\includegraphics[width=3em,height=0.6em]{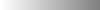}~high) for various GP priors (infinite-width networks), and the bottom row shows finite-width MLP neural network results (one hidden layer, 50 nodes, MC dropout) with the network activation function matching the GP prior on the top row.
  }  
  \label{fig:teaser}
  \vspace*{-1em}
\end{figure}

Along the line of studying random networks, \citet{Neal:1995} showed that under certain assumptions, random neural networks with one hidden layer converge to a Gaussian process (GP, \cite{Rasmussen+Williams:2006}) in the limit of infinite width. Since then, explicit links between common neural network activation functions and GP covariance (kernel) functions have been shown (\eg, ERF and RBF activations by \cite{williams97computing} and ReLU and step activations by \cite{cho+saul:2009}, see \cref{fig:teaser}). Recently, this equivalence was extended to deep networks \cite{matthews2018gaussian,lee2018deep}. \citet{Gal+Ghahramani:2016} leveraged the connection for approximate variational inference in neural networks. The call for more work on deep net priors (\eg, \cite{pearce2019expressive,Sun+Zhang+Shi:2019,flam2017mapping}) is also motivated by  empirical findings \cite{Wenzel+Roth+Veeling:2020}, and we recognize two separate problems in deep learning: {\em (i)}~designing network architectures that support our domain knowledge of the modelling task (prior assumptions), and {\em (ii)}~performing inference and learning with the model. Recent interest in Bayesian deep learning has been in {\em (ii)} (see, \eg, \cite{hernandez2015probabilistic,blundell2015weight,maddox19_SWAG,ritter2018a_kfac_laplace}), while we tackle {\em (i)}, and choose to apply a simple and conservative inference method---we resort to Monte Carlo (MC) dropout in our experiments.

Orthogonal to deep learning, in support vector machines \cite{cortes1995support}, kernel methods \cite{hofmann2008kernel}, spatial statistics \cite{Cressie:1991}, and GP models,  the focus is on the choice and crafting of a {\em kernel} (covariance/covariogram function). The kernel captures prior assumptions related to the model functions such as continuity, differentiability, periodicity, invariances, \etc\ Stationarity (translation invariance of the kernel) is often a sought-after property in these models as it induces the behaviour of functions reverting back to the prior outside informative regions in the problem domain (decision boundaries/data samples). Arguably the most used GP kernel is the Mat\'ern class \cite{Matern:1960,Rasmussen+Williams:2006}, which features stationary kernels with continuous sample functions of various degree of smoothness. This class has the RBF (squared exponential or Gaussian) and the exponential (Ornstein--Uhlenbeck) covariance functions as limiting cases of infinite smoothness and non-differentiable sample paths, respectively (see, \eg, \cite{Rasmussen+Williams:2006}). 

Previous approaches have sought to map various activation functions to their kernel counterparts. We take the opposite approach by introducing a new family of non-linear neural network activation functions that mimic the widely-used Mat\'ern family in GP models. The derivation is made possible by rekindling the link between classical control theory and neural networks. The class we propose spans a range of locally stationary models of various degrees of mean-square differentiability. We show an explicit link to the corresponding GP models in the case of one infinitely wide hidden layer. These activation functions result in similar appealing properties to their counterparts in GP models, and we demonstrate that the local stationarity property together with limited mean-square differentiability shows good performance and uncertainty quantification in Bayesian deep learning tasks. In particular, the local stationarity can help in tackling overconfidence in out-of-distribution detection.

\begin{figure}[!t]
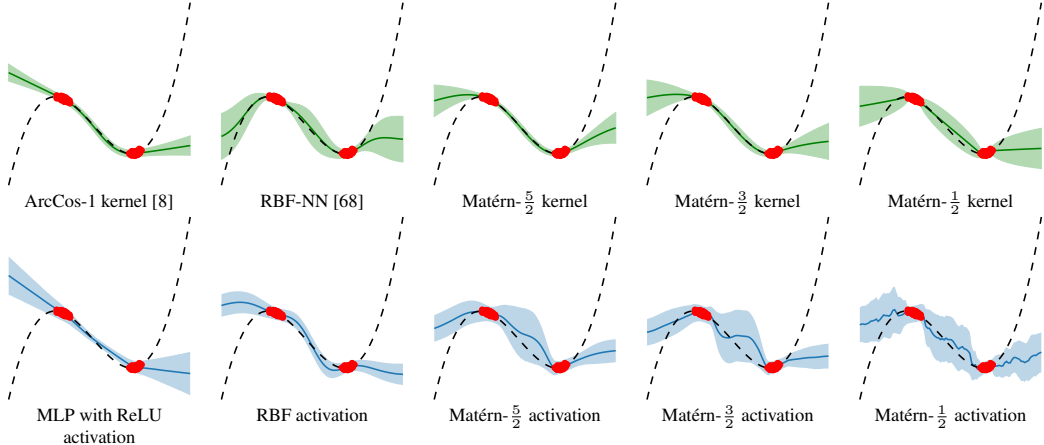

  \scriptsize
  \pgfplotsset{hide axis,scale only axis,width=\figurewidth,height=\figureheight}
  \setlength{\figurewidth}{.19\textwidth}
  \setlength{\figureheight}{\figurewidth}  
  \begin{subfigure}[t]{.19\textwidth}
    \centering
    \input{./fig/1D_GP_arcosine1.tex}\\[0em]
    ArcCos-1 kernel \cite{cho+saul:2009}
  \end{subfigure}
  \hfill
  \begin{subfigure}[t]{.19\textwidth}
    \centering
    \input{./fig/1D_GP_rbf.tex}\\[0em]
    RBF-NN \cite{williams97computing}
  \end{subfigure}
  \hfill
  \begin{subfigure}[t]{.19\textwidth}
    \centering
    \input{./fig/1D_GP_matern52.tex}\\[0em]
    Mat\'ern-$\frac{5}{2}$ kernel
  \end{subfigure}
  \hfill
  \begin{subfigure}[t]{.19\textwidth}
    \centering
    \input{./fig/1D_GP_matern32.tex}\\[0em]
    Mat\'ern-$\frac{3}{2}$ kernel
  \end{subfigure}
  \hfill
  \begin{subfigure}[t]{.19\textwidth}
    \centering
    \input{./fig/1D_GP_matern12.tex}\\[0em]
    Mat\'ern-$\frac{1}{2}$ kernel    
  \end{subfigure}  
  \\
  \begin{subfigure}[t]{.19\textwidth}
    \centering
    \input{./fig/1D_MCdrop_relu.tex}\\[0em]
    MLP with ReLU activation
  \end{subfigure}
  \hfill
  \begin{subfigure}[t]{.19\textwidth}
    \centering
    \input{./fig/1D_MCdrop_rbf.tex}\\[0em]
    RBF activation
  \end{subfigure}
  \hfill
  \begin{subfigure}[t]{.19\textwidth}
    \centering
    \input{./fig/1D_MCdrop_matern52.tex}\\[0em]
    Mat\'ern-$\frac{5}{2}$ activation
  \end{subfigure}
  \hfill  
  \begin{subfigure}[t]{.19\textwidth}
    \centering
    \input{./fig/1D_MCdrop_matern32.tex}\\[0em]
    Mat\'ern-$\frac{3}{2}$ activation
  \end{subfigure}
  \hfill
  \begin{subfigure}[t]{.19\textwidth}
    \centering
    \input{./fig/1D_MCdrop_matern12.tex}\\[0em]
    Mat\'ern-$\frac{1}{2}$ activation
  \end{subfigure}
  \caption{Illustrative examples of 1D regression tasks (shading covering two sigma) with GP models on the top row and neural networks (50 hidden units, MC dropout) on the bottom. Our proposed models capture similar behaviour as their GP counterparts, with `stiffness' reducing from left to right. The predictive variance scale for the neural network models is strongly affected by the choice of hyperparameters, such as the dropout rate, and the focus of this example is showing how the variance around the training data relates to the variance away from training points.}  
  \label{fig:1d}
  \vspace{-1em}
\end{figure}

\subsection{Related Work}
We naturally desire predictive models to know what they do not know: that is, to have well-calibrated out-of-sample or \emph{out-of-distribution (OOD)} detection. This is often of vital concern for safety-critical applications \cite{amodei16_concrete}. However, overly confident OOD predictions is a well-recognized problem in state-of-the-art discriminative neural networks \cite{goodfellow15explaining,nguyen2015deep,hendrycks17baseline,liang2018enhancing} and deep generative models \cite{hendrycks2018deep,nalisnick2018do,choi18_waic_but_why,ren19_likel_ratios_out_of_distr_detec}. To clarify our terminology, we consider `OOD detection' to be a broad problem that includes `anomaly detection'/`outlier detection' \cite{choi18_waic_but_why,hendrycks2018deep}), `open-set recognition' \cite{dhamija18_reduc_networ_agnos,Boult19}, and robustness to `domain shift', `data set shift', and `covariate shift' \cite{bradshaw17_adver_examp_uncer_trans_testin,Snoek+Ovadia19_trust} or `rubbish class' \cite{goodfellow15explaining} and `fooling' \cite{nguyen2015deep} examples.
OOD data can refer to unseen classes from the same data set (\eg, unknown bacteria \cite{ren19_likel_ratios_out_of_distr_detec} or fraudulent credit card transactions \cite{choi18_waic_but_why}), or a different data set (\eg, \textsc{NotMNIST} from \textsc{MNIST} \cite{Lakshminarayanan17_deep_ensembles,ritter2018a_kfac_laplace}, \textsc{SVHN} from \textsc{CIFAR-10} \cite{nalisnick2018do}). Data can also be OOD by construction: `rubbish class' or `fooling' examples are synthetic inputs, \eg, random noise images \cite{nguyen2015deep}. OOD inputs are generally \emph{far away} from training data, whereas an arguably related problem arises from minor perturbations from the training data, such as common corruptions \citep{hendrycks2018benchmarking} and adversarial examples \citep{szegedy14_intriguing_properties}.

For many classifier neural networks, part of the problem is intrinsic to their very architecture. Goodfellow~\etal~\cite{goodfellow15explaining} attributed overconfidence on far-away `rubbish' examples to the linear behaviour of ReLU networks in high dimensions, while RBF networks were shown to be immune. This was later formalized, in that all networks with piecewise affine activations (\eg, ReLU, leaky ReLU) lead to continuous piecewise affine classifiers \cite{arora2018understanding,croce2018randomized}, which produce predictions with arbitrarily high confidences on far-away data, whereas RBF networks have almost-uniform confidence \cite{hein2019relu}. While many OOD detection methods rely on specific assumptions or problem formulations (such as knowledge of the OOD distribution \cite{dhamija18_reduc_networ_agnos,hein2019relu}, or augmenting classifier with a reject class (see \cite{dhamija18_reduc_networ_agnos} background section for examples)), in our work we directly address the neural network prior itself, which is largely governed by its \emph{activation functions}.

In contrast to neural networks, GPs with stationary kernels do not overconfidently extrapolate far from the training data, much like RBF networks. However, they have limited representational power, but would perhaps be more widely applied if not for their associated computational complexity. As such, we use the previously mentioned GP-neural network correspondence to incorporate the desirable OOD behaviour of Mat\'ern kernel GPs into neural networks, via a new class of activation functions. Related work addressing the modelling problem in deep learning has similarly revolved around designing functional priors \cite{Sun+Zhang+Shi:2019,flam2017mapping}, such as through BNN architectural choices \cite{pearce2019expressive}. However, the focus has not been on improving OOD behaviour, with the exception of noise contrastive priors (NCPs, \cite{hafner18_noise_contr_prior_funct_uncer}) where, however, generating noisy samples at data boundaries can be hard in practice.

Our approach of studying stationary kernels through their spectral density is related to a wide range of previous work in Gaussian process models. This duality has given rise to leveraging Fourier features (see, \eg, \cite{rahimi2008,lazaro2010sparse,hensman2018variational,solin2020hilbert}) by projecting the GP problem on a set of harmonic basis functions. While we share the idea of using the Fourier duality, the resulting model is spanned by different basis functions. Conventional sinusoidal Fourier features  enforce (global) stationarity with the approximation based on \cref{eq:duality}, while our approach is locally stationary as defined by \cref{eq:covfun-nn}. This same global vs.\ local difference is also shared between this work and the SIREN activations of \cite{sitzmann2020implicit}.

\section{Random Feedforward Networks and Gaussian Processes}
Gaussian process (GP, \cite{Rasmussen+Williams:2006}) models admit the form of a GP prior $f(\vx) \sim \GP(\mu(\vx), \kappa(\vx,\vx'))$ and a likelihood (observation) model $\vy \mid \vf\sim\prod_{i=1}^{n} p(y_{i} \mid f(\vx_{i}))$, where the data $\mathcal{D} = \{(\vx_i, y_i)\}_{i=1}^n$ are input--output pairs, $\mu(\vx)$ the mean function, and $\kappa(\vx,\vx')$ the covariance function of the GP prior. This probabilistic machine learning paradigm covers many standard modelling problems, including regression and classification tasks, where GPs not only predict well but also enable uncertainty estimation and model selection via the marginal likelihood.
The Gaussian process is completely specified by its mean and covariance function, which encapsulate the assumptions about the sample processes $f$ (such as continuity, differentiability, periodicity, \etc):
\begin{equation}
  \mu(\vx) := \mathrm{E}[f(\vx)] \qquad \text{and} \qquad 
  \kappa(\vx,\vx') := \mathrm{E}[(f(\vx)-\mu(\vx))(f(\vx)-\mu(\vx))^*].
\end{equation}
Without loss of generality, we limit our interest to zero-mean ($\mu(\vx):=0$) GP priors. In practice implementations work with an $n {\times} n$ Gram (covariance) matrix $\MK$ with $\kappa(\vx_i,\vx_j)$ as the $ij$\textsuperscript{th} entry for $\forall\vx \in \mathcal{D}$. This gives rise to a prohibitive cubic scaling $\mathcal{O}(n^3)$ in the number of data due to associated matrix inversions, as well as the non-parametric nature of GPs, where the number of parameters in the model is not fixed, but rather spanned by the number of data points.

The link between feedforward neural networks (NN) and GPs is generally well understood \cite{matthews2018gaussian}. \citet{Neal:1995} showed that a random (untrained) single-layer network converges to a GP in the limit of infinite width.  Let $\sigma(\cdot)$ be some non-linear (activation) function such as the ReLU or sigmoid, and $\vw$ and $b$ be the network weights and biases. We can define the associated kernel for the infinite-width network (under assumptions on Gaussian weights) to be formulated in terms of \cite{Neal:1995}
\begin{equation}\label{eq:covfun-nn}
  \kappa(\vx,\vx') = \int p(\vw)\,p(b)\,\sigma(\vw\T\vx+b)\,\sigma(\vw\T\vx'+b)\,\dd\vw \dd b,
\end{equation}
where $p(\vw)$ is a multivariate distribution and $p(b)$ is a univariate Gaussian. In the absence of a closed-form solution, a Monte Carlo approximation to the above integral arrives at a random or rank-deficient (see \cite{Gal+Ghahramani:2016}) formulation for the covariance function
\begin{equation}\label{eq:covfun-nn-mc}
  \hat{\kappa}(\vx,\vx') = \frac{1}{K} \sum_{k=1}^K \sigma(\vw\T\vx+b)\,\sigma(\vw\T\vx'+b)
\end{equation}
with $\vw \sim p(\vw)$ and $b \sim p(b)$, and $K$ having the interpretation of the number of hidden units in a single hidden layer NN approximation. 
This representation has links to {\em Mercer's theorem} \cite{mercer1909functions} which states that any positive-definite kernel can be represented as the inner product between a fixed set of features, evaluated at $\vx$ and $\vx'$: $\kappa(\vx,\vx') = \vh(\vx)\T\vh(\vx')$.
From the formulation in \cref{eq:covfun-nn} it is easy to see that for a given $\sigma(\cdot)$ the corresponding kernel can be recovered by solving this integral either by sampling as in \cref{eq:covfun-nn-mc} or in closed form as has been done, \eg, for the RBF and ERF \cite{williams97computing}, ReLU and step \cite{cho+saul:2009}, leaky ReLU \cite{tsuchida2018invariance}, and cosine activations \cite{pearce2019expressive} (\cf\ \cref{fig:teaser}). However, solving the {\em inverse problem} of recovering $\sigma(\cdot)$ given $\kappa(\cdot,\cdot)$ is typically harder and has received less attention.

\section{Methods}
We derive activation functions $\sigma(\cdot)$ (non-coincidentally called {\em transfer functions} in 1980s and `90s neural networks literature) resulting in random networks that capture the behaviour induced by the Mat\'ern class \cite{Matern:1960,Rasmussen+Williams:2006}. We start from a fully stationary setting, which we then relax to local stationarity in \cref{sec:local-stationarity} in order to match the implicit input density assumption in random networks (giving our main result). We conclude this section by providing an alternative view through functional analysis.

\subsection{Transfer Function Approach to Activation}
\label{sec:transfer}
A {\em stationary} (homogeneous) covariance function is invariant to translations of the input space. This means that the covariance structure of the model functions $f(\vect{x})$ is the same regardless of the absolute position in the input space, and thus the covariance function can be parameterized as $\kappa(\vx,\vx') \triangleq \kappa(\vx-\vx') = \kappa(\vr)$. For stationary GP priors, the covariance function can be written equivalently in terms of its spectral density function. This results from \emph{Bochner's theorem} (see, \eg, \cite{Akhiezer+Glazman:1993, DaPrato:1992}) which states a bounded continuous positive definite function $\kappa(\vr)$ can be represented as
\begin{equation}
\kappa(\vr) = \frac{1}{(2\pi)^{d}} \int 
       \exp\left(\imag \, \vomega\T \vect{r}\right) \, \mu(\!\dd \vomega),
\end{equation}
where $\mu$ is a positive measure and $\vr \in \mathbb{R}^d$. If the measure $\mu(\vomega)$ has a density, it is called the \emph{spectral density} $S(\vectb{\omega})$ corresponding to the covariance function $\kappa(\vect{r})$. This is the Fourier duality of covariance and spectral density, which is known as the \emph{Wiener--Khinchin theorem} (see, \eg, \cite{Rasmussen+Williams:2006}):
\begin{equation} \label{eq:duality}
  \kappa(\vr) = \frac{1}{(2\pi)^d} \int S(\vomega) \,
       \exp( \imag \, \vomega\T \vect{r}) \dd \vomega
  \quad \text{and} \quad
  S(\vomega) = \int \kappa(\vr) \,
       \exp( -\imag \, \vomega\T \vr) \dd \vr.
\end{equation}
If $d>1$, these identities show that if the covariance function is \emph{isotropic}---that is, it only depends on the Euclidean norm $\norm{\vr}$ such that $\kappa(\vect{r}) \triangleq \kappa(\norm{\vr})$ (thus being invariant to all rigid motions of the input space)---then the spectral density will also only depend on the norm of the dual input variable $\vomega$. In one-dimension these definitions coincide, and in the following we restrict our interest to 1D projections in the spirit of \cref{eq:covfun-nn}.
From the above, we can also define the corresponding (power) spectral density of the process, which is the square of the absolute value of the Fourier transform of the process. If we denote the spectral density of white noise $|W(\imag \, \omega)|^2 = q^2$, the spectral density of the process can be decomposed as (see \cite{Glad+Ljung:2000} for an overview on the concepts in control theory)
\begin{equation} \label{eq:specfact}
  S(\omega) = q^2 \, |G(\imag \, \omega)|^2 = {G(\imag \, \omega)} \, q^2 \, G(-\imag \, \omega),
\end{equation}
where $G(\cdot)$ is typically referred to as the {\em transfer function} in signal processing. A transfer function corresponds to a stable system if and only if all of its poles (zeros of the denominator) are in the upper half of the complex plane. The (often non-trivial) procedure for finding a {(stable) transfer function $G(\imag\,\omega)$} is called spectral factorization. 

An implication of \cref{eq:duality,eq:specfact} is that, if we take the Fourier transform of $f \sim \GP(0,\kappa(r))$ and solve the Fourier transform of the process $F(\imag \, \omega)$, we get
  $F(\imag \, \omega) = G(\imag \, \omega) \, W(\imag \, \omega)$,
where $W(\imag \, \omega)$ is the (formal) Fourier transform of the white noise. This equation can be interpreted such that the process $F(\imag \, \omega)$ is obtained by feeding white noise through a system with the transfer function $G(\imag \, \omega)$ (see \cite{Sarkka+Solin+Hartikainen:2013} for discussion on relation to temporal GPs). This takes an interesting role under the analysis of random networks in the sense that this relation is exactly what we want to capture in terms of the finding $\sigma(\cdot)$ w.r.t.\ \cref{eq:covfun-nn} (feeding white-noise through the system formally corresponds to a network with Gaussian weights). Thus we can reduce the problem to analysing the transfer function $G(\imag\,\omega)$.

\begin{figure}[!t]
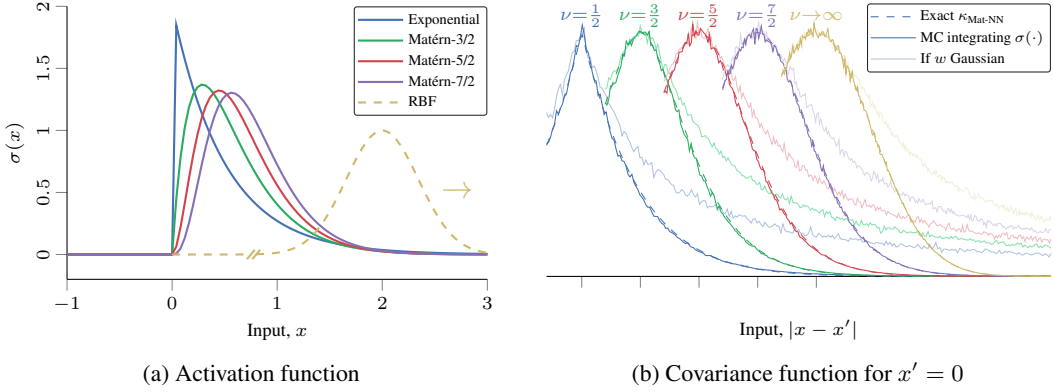

  \centering
  \scriptsize
  \pgfplotsset{scale only axis,width=\figurewidth,height=\figureheight,xtick align=outside,ytick align=outside}
  \pgfplotsset{legend style={inner xsep=1pt, inner ysep=1pt, row sep=0pt},legend style={at={(1.00,1.00)},anchor=north east},legend style={rounded corners=1pt},y tick label style={rotate=90}, legend style={nodes={scale=0.8, transform shape}}} 
  \setlength{\figurewidth}{.4\textwidth}
  \setlength{\figureheight}{.65\figurewidth}  
  \begin{subfigure}[b]{.48\textwidth}
    \input{./fig/activation.tex}
    \caption{Activation function}
    \label{fig:activation}
  \end{subfigure}
  \hfill  
  \begin{subfigure}[b]{.48\textwidth}
    \pgfplotsset{hide y axis,xtick align=outside}
    \setlength{\figurewidth}{\textwidth}
    \input{./fig/covfun.tex}
    \caption{Covariance function for $x'=0$}
    \label{fig:covfun}
  \end{subfigure}\\[-3pt]
  \caption{(a)~Activation functions $\sigma(\cdot)$ corresponding to the Mat\'ern kernel class for various typical degrees of smoothness $\nu$. (b)~Corresponding covariance functions calculated by MC integration and compared to their exact dashed counterparts (peaks shifted for clarity). Light curves are the same under Gaussian weights instead of binary white weights. When $\nu{\to}\infty$, we recover the RBF-NN \cite{williams97computing}.}
  \label{fig:cov-act}
  \vspace{-1em}
\end{figure}

The rest of this section is concerned with the particular case of the stationary Mat\'ern family \cite{Stein:1999}:
\begin{equation}\label{eq:matern}
  \kappa_\mathrm{Mat.}(\vx,\vx') = \frac{2^{1-\nu}}{\Gamma(\nu)}\left(\sqrt{2\nu}\,\frac{\|\vx-\vx'\|}{\ell}\right)^\nu \mathrm{K}_\nu\!\left(\sqrt{2\nu}\,\frac{\|\vx-\vx'\|}{\ell}\right), 
\end{equation}
where $\nu$ is a smoothness and $\ell$ a characteristic length-scale parameter, $\mathrm{K}_\nu(\cdot)$ the modified Bessel function, and $\Gamma(\cdot)$ the gamma function. For the Mat\'ern class, the processes $f(\vx)$ are $\lceil \nu \rceil -1$ times mean-square differentiable. In 1D, under the {\em angular frequency} convention (by \cref{eq:duality}), we realize the spectral density function is of the following factorizable form (derivation in \cref{app:matern-transfer}):
\begin{equation}
  S(\omega) \propto \left( \lambda^2+\omega^2 \right)^{-(\nu+1/2)}
    = { \left( \lambda + \imag\,\omega \right)^{-(\nu+1/2)}} \, 
    \left( \lambda- \imag\,\omega \right)^{-(\nu+1/2)},
\end{equation}
where $\lambda=\sqrt{2\nu}/\ell$. Following \cref{eq:specfact}, we can collect the transfer function of the corresponding stable part as
  $G(\imag \, \omega) = \left( \lambda+ \imag\,\omega \right)^{-(\nu+1/2)}$.
The remaining (power) spectral density (formally that of the driving white noise process $w(t)$ that captures the scaling coefficients) is
$q^2 = {2 \pi^{1/2} \lambda^{2\nu}\Gamma(\nu+1/2)}/{\Gamma(\nu)}$.
Now taking the inverse Laplace transform of the transfer function $G(\imag \, \omega)$ gives (this can be shown by considering the simplified expression $\Laplace[x^{\alpha-1}\,\exp(-a\,x)](s) = \Gamma(\alpha)\,(s+a)^{-\alpha}$, for $\alpha>0$, details in \cref{app:matern-activation})
and using the standard property of expanding the transform to the real line, gives us the transfer (or {\em activation}) in the input space:
\begin{equation}\label{eq:activation}
  \sigma(x)  = \frac{q}{\Gamma(\nu+1/2)}\,\Theta(x)\,x^{\nu-1/2}\,\exp(-\lambda\,x),
\end{equation}
where $\Theta(\cdot)$ is the Heaviside step function.
For $\nu>1$, these non-linear functions are smooth, continuous, and continuously differentiable. For $\nu<1$, the functions are not smooth (rather taking the interesting form of an exponentially decaying step function), agreeing with the properties of the Mat\'ern family. The function is not monotonic, which will be discussed in later sections. \cref{fig:cov-act} shows realizations of $\sigma(\cdot)$ for various degrees of smoothness $\nu$. In terms of activation, the scaling can be made arbitrary, but scaling coefficients as in \cref{eq:activation} can help with stabilizing training.
With $\nu \to \infty$ \cref{eq:activation} approaches a bell curve (see \cref{app:limit-rbf} for a proof), recovering the relation between the Mat\'ern family and the RBF kernel---giving the result of \cite{williams97computing} (included in \cref{fig:activation,fig:covfun}).

\subsection{Local Stationarity from Modulating Envelopes}
\label{sec:local-stationarity}
Considering the activation function we have in \cref{eq:activation} under the random network formalism of \cref{eq:covfun-nn}, we observe that this covariance function cannot be stationary due to the distributions of the weights being centered around zero, and hence no translational symmetry is present. Instead, the covariance is locally stationary (see discussion in \cite{genton2001classes}) in the same sense as the RBF-NN covariance function (see \cite{williams1998computation,Rasmussen+Williams:2006}). This means that for any finite value of $\sigma_\mathrm{b}^2$ in $p(b) = \mathrm{N}(b \mid 0,\sigma_\mathrm{b}^2)$ and $p(\vw)$ being binary white, we can write a composite covariance function with a Gaussian decay envelope
\begin{equation}\label{eq:Mat-NN}
\kappa_\textrm{Mat-NN}(\vx,\vx') \propto \exp(-\vx{}\T\vx/2\sigma_\mathrm{m}^2)\,\kappa_\textrm{Mat.}(\vx,\vx')\exp(-{\vx'}{}\T{\vx'}/2\sigma_\mathrm{m}^2),
\end{equation}
where $\sigma_\mathrm{m}^2 = 2\sigma_\mathrm{b}^2 + \ell^2$.
As an example (and sanity check), \cref{fig:covfun} shows that we recover the exact covariance function $\kappa_\textrm{Mat-NN}(\cdot,\cdot)$ (dashed, barely visible due to almost exact match) using our proposed $\sigma(x)$ by MC integration in \cref{eq:covfun-nn-mc} (we use $K=10{,}000$). For completeness, we also show the curves resulting from unit Gaussian weights on $p(\vw)$ (light colours).

\subsection{Alternative View Through Green's Function}
\label{sec:green}
In the case of the isotropic RBF kernel, the network output is a linear combination of radial basis functions, as observed by \citet{broomhead1988multivariable} (see \cite{wahba1990spline,williams1998computation} for discussion on the role of radial basis functions). This functional analysis view is taken by \citet{poggio1990networks} in deriving RBF networks. A similar construction could be applied here, but for the Mat\'ern family this becomes unpractical, as explained below.

For any covariance function $\kappa(\vx,\vx')$, we can define the associated covariance operator $\op{K}$ as follows:
  $\op{K} \phi = \int \kappa(\cdot,\vx') \, \phi(\vx') \dd\vx'$.
For stationary covariance functions, this can also be written as a convolution. 
By assuming the inputs $\vx$ to obey some density (see \cite{Williams+Seeger:2000}), we  consider an inner product defined by that density. Let the inner product be defined as
$ \innerp{f,g} = \int f(\vx)\,g(\vx)\,w(\vx) \dd \vx, $
where $w(\vx)$ is some positive weight function such that $\int w(\vx) \dd \vx < \infty$. In terms of this inner product, we define the operator 
  $\op{K}f = \int \kappa(\cdot,\vx) \, f(\vx) \, w(\vx) \dd \vx$.
This operator is self-adjoint with respect to the inner product, $\innerp{\op{K}f,g} = \innerp{f,\op{K}g}$, and according to the spectral theorem there exists an orthonormal (in sense of $\int \varphi_i(\vx)\,\varphi_j(\vx)\,w(\vx) \dd \vx = \delta_{ij}$) set of basis functions and positive constants, $\{(\varphi_j(\vx), \gamma_j)\}$, that satisfies the eigenvalue equation
  $(\op{K} \varphi_j)(\vx) = \gamma_j \, \varphi_j(\vx)$.
Thus $\kappa(\vx,\vx')$ has the series expansion (\cf\ Hilbert--Schmidt theorem) 
  $\kappa(\vx,\vx') = \sum_{j=1}^\infty \gamma_j \, \varphi_j(\vx) \, \varphi_j(\vx')$.

In the case of the RBF kernel and assuming a Gaussian input density $w(\vx)$, these eigenvalues and eigenfunctions are available in closed form (see \cref{app:closed_form_RBF}). This directly relates to the envelope in the RBF-NN kernel, which takes the form $w^{1/2}(\vx)$ (see, \cite{williams1998computation}, for a brief discussion). The relation to $\sigma(\cdot)$ is best understood through the so called Green's function $G(\cdot,\cdot)$ of the operator (the symbol re-used here on purpose), which is here defined through the property: $\op{K}\, G(\vx,\vx') = \delta(\vx-\vx')$ (for stationary systems, $G(\vx,\vx') \triangleq G(\vx-\vx')$).
From the definition it is apparent that $G(\cdot)$ generalizes the concept of impulse response in system theory to linear operators. Recalling that the transfer function is the Laplace transform of the impulse response, we see the relation between $\sigma(\cdot)$, Green's function, and \cref{sec:transfer}. Solving the Green's function associated with $\op{K}$ (and thus $\kappa(\cdot,\cdot)$) is non-trivial, and perhaps best grasped through the relation $G(\vx,\vx') = \sum_{j=1}^\infty \gamma_j^{-1}\varphi^*(\vx)\,\varphi(\vx')$ (see, \eg, \cite{wahba1990spline}). This relation helps theoretical understanding, but for deriving the activation function for, \eg, the Mat\'ern kernel this is highly impractical. In theory, one could resort to numerical approximations, but recovering $G(\cdot,\cdot)$ or $\sigma(\cdot)$ this way becomes unstable due to the inversion of the eigenvalues.

\section{Experiments}
We have included a comprehensive set of experiments that concentrate on analysing the quality of model prediction uncertainty and OOD tests. We consider illustrative toy data sets and standard classification benchmarks, analyse image classification, and finally provide a realistic safety-critical application example in radar emitter classification.
The experiments were implemented in GPflow \cite{GPflow:2017} (GPs and GPDNN), GPyTorch \cite{GPyTorch} (SV-DKL), and the rest in PyTorch (see \cref{app:experiments}). For all neural network models using the Mat\'ern activation functions the length-scale parameter $\ell$ is fixed as the preceding layer(s) take care of scaling the inputs, which serves the same purpose. \cref{app:experiments} lists full details of all the experiments.

\begin{table}[t!]
  \caption{Examples of UCI classification tasks, showing the Mat\'ern-3/2 NN directly gives competitive accuracy, mean negative log predictive density (NLPD), and area under receiver operating characteristic curve (AUC) to sparse GPs or NN+GP hybrids. More results in \cref{app:benchmark}.}
  \label{tbl:benchmarks}
  \scriptsize
  \tiny

  \setlength{\tabcolsep}{0pt}
  \setlength{\tblw}{0.09\textwidth}  
  \begin{tabularx}{\textwidth}{l @{\extracolsep{\fill}} c @{\extracolsep{\fill}} c @{\extracolsep{\fill}} c @{\extracolsep{\fill}} C{\tblw} C{\tblw}  C{\tblw} C{\tblw}  C{\tblw} C{\tblw} C{\tblw} C{\tblw}}
  \toprule
  \multicolumn{4}{l}{(10-fold cv)} & \multicolumn{4}{c}{NLPD} & \multicolumn{4}{c}{ACC} \\
  & $n$ & $d$ & $c$ & SVGP & GPDNN & SV-DKL & Mat\'ern act. & SVGP & GPDNN & SV-DKL & Mat\'ern act. \\
  \midrule
  Adult & 45222 & 14 & 2 & $.344{\pm}.006$ & $.435{\pm}.014$ & $\bf.316{\pm}.006$ & $\bf.316{\pm}.007$ & $.842{\pm}.005$ & $.821{\pm}.037$ & $\bf.855{\pm}.004$ &  $.854{\pm}.005$\\
  Connect-4 & 67556 & 42 & 3 & $.629{\pm}.010$ & $.763{\pm}.018$ & $.459{\pm}.016$ & $\bf.450{\pm}.008$ & $.750{\pm}.006$ & $.768{\pm}.006$ & $.827{\pm}.009$ &  $\bf.828{\pm}.004$\\ 
  Covtype & 581912 & 54 & 7 & $.494{\pm}.002$ & $.722{\pm}.025$ & $\bf.101{\pm}.005$ & $.118{\pm}.003$ & $.787{\pm}.002$ & $.842{\pm}.008$ & $\bf.962{\pm}.001$ &  $.958{\pm}.001$\\ 
  Diabetes & 768 & 8 & 2 & $.506{\pm}.034$ & $.634{\pm}.012$ & $.691{\pm}.005$ & $\bf .486{\pm}.081$ & $.759{\pm}.056$ & $.744{\pm}.040$ & $.507{\pm}.143$ &  $\bf.766{\pm}.044$\\ 
  \midrule
  \multicolumn{4}{l}{} & \multicolumn{4}{c}{AUC} & \multicolumn{4}{c}{}\\
  & $n$ & $d$ & $c$ & SVGP & GPDNN & SV-DKL & Mat\'ern act. & \multicolumn{4}{c}{}\\
  \cmidrule{0-7}
  Adult & 45222 & 14 & 2  & $.893{\pm}.004$ & $.774{\pm}.052$ & $.912{\pm}.003$ &  $\bf.913{\pm}.004$ & \multicolumn{4}{c}{}\\
  Connect-4 & 67556 & 42 & 3  & $.824{\pm}.005$ & $.675{\pm}.019$ & $.909{\pm}.013$ &  $\bf.913{\pm}.004$ & \multicolumn{4}{c}{}\\  
  Covtype & 581912 & 54 & 7  & $.971{\pm}.001$ & $.943{\pm}.015$ & $\bf.998{\pm}.000$ &  $\bf.998{\pm}.000$ & \multicolumn{4}{c}{}\\  
  Diabetes & 768 & 8 & 2  & $.817{\pm}.049$ & $.769{\pm}.053$ & $.512{\pm}.095$ & $\bf.838{\pm}.051$ & \multicolumn{4}{c}{}\\  
  \cmidrule[.5pt]{0-7}
  \end{tabularx}
  \vspace*{-1em}
\end{table}

\paragraph{Illustrative toy examples}
In \cref{fig:teaser}, we consider the binary Banana classification tasks under the presence of various GP priors. The inference is performed both with a variational GP (VGP, \cite{GPflow:2017}) and using a small-size NN (to show noisiness) with one hidden layer (50 hidden units and activations corresponding to the top-row GP priors). Low predictive variance outside the training data is present in all methods, but the RBF and Mat\'ern activations show localization around the training data.
\cref{fig:1d} shows the typical Bayesian deep learning toy regression task for in-between uncertainty prediction. Our activations capture similar behaviour as their GP counterparts, with `stiffness' reducing with $\nu$. 

\paragraph{Benchmark classification tasks} 
In \cref{tbl:benchmarks}, we consider UCI benchmark classification tasks (including one small-data example) where we compare classification accuracy and negative log predictive density (NLPD) that penalizes both misclassification and miscalibrated uncertainty. We compare our model to NN+GP hybrids GPDNN \cite{bradshaw17_adver_examp_uncer_trans_testin} and SV-DKL \cite{Wilson+Hu+Salakhutdinov:2016}, and a sparse GP method (SVGP, \cite{Hensman+Matthews+Ghahramani:2015}). We use a Mat\'ern $\nu=3/2$ covariance function for all GPs and the corresponding activation. The NN architectures in all methods are the same (a fully connected network with layers d-1000-1000-500-50-c). We only consider the Mat\'ern activation here, as, \eg, the ReLU consistently performs poorly in terms of NLPD. The Mat\'ern activation results are consistent and agree or outperform the hybrid methods that need to balance between training the NN and GP (which can be unstable and require approximate GP method in practice), when essentially having the same prior as our method. Details and further results for other activation functions can be found in \cref{app:benchmark}. We also include results for the SIREN \cite{sitzmann2020implicit} activation function, which induces a non-local infinitely smooth prior, comparable to the RBF.

\begin{figure}[!t]
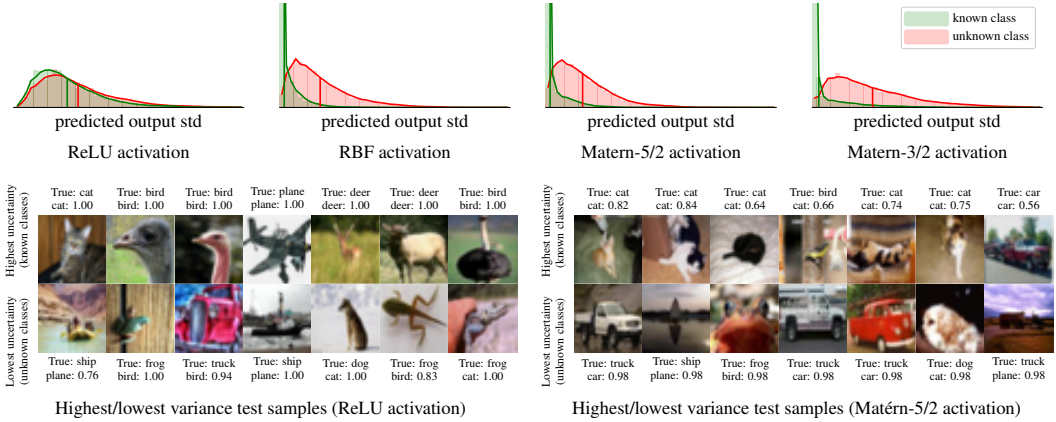

  \vspace*{-1em}
  \scriptsize
  \pgfplotsset{axis on top,scale only axis,width=\figurewidth,height=\figureheight}
  \pgfplotsset{legend style={inner xsep=1pt, inner ysep=1pt, row sep=0pt},legend style={rounded corners=1pt},legend style={nodes={scale=1.0, transform shape}}}
  \setlength{\figurewidth}{.22\textwidth}
  \setlength{\figureheight}{.45\figurewidth}
  \begin{subfigure}[t]{.24\textwidth}
    \centering
    \input{./fig/cifar/cifar10_relu.tex}\\[0em]
    ReLU activation
  \end{subfigure}
  \hfill
  \begin{subfigure}[t]{.24\textwidth}
    \centering
    \input{./fig/cifar/cifar10_rbf.tex}\\[0em]
    RBF activation
  \end{subfigure}
  \hfill
  \begin{subfigure}[t]{.24\textwidth}
    \centering
    \input{./fig/cifar/cifar10_matern52.tex}\\[0em]
    Matern-5/2 activation
  \end{subfigure}
  \hfill
  \begin{subfigure}[t]{.24\textwidth}
    \centering
    \input{./fig/cifar/cifar10_matern32.tex}\\[0em]
    Matern-3/2 activation
  \end{subfigure}\\[1em]
  \begin{subfigure}[t]{.49\textwidth}
    \centering
    \setlength{\figurewidth}{.18\textwidth}
    \setlength{\figureheight}{\figurewidth}
    \resizebox{\textwidth}{!}{
    \begin{tikzpicture}[inner sep=0]

      \tiny

      \node[rotate=90,text width=1.5\figureheight,align=center] at (.25,.35) {\tiny Highest uncertainty (known classes)};
      \node[rotate=90,text width=1.5\figureheight,align=center] at (.25,-1.65) {\tiny Lowest uncertainty (unknown classes)};

      \foreach \x [count=\i] in {True: cat\\cat: 1.00,True: bird\\bird: 1.00,True: bird\\bird: 1.00,True: plane\\plane: 1.00,True: deer\\deer: 1.00,True: deer\\deer: 1.00,True: bird\\bird: 1.00}
        \node[align=center]
          (\i) at ({\figurewidth*\i},{\figureheight*0.75})
          {\x};

      \foreach \x [count=\i] in {0,1,2,3,4,5,6}
        \node[draw=white,fill=black!20,minimum size=\figurewidth,inner sep=0pt]
          (\i) at ({\figurewidth*\i},{-\figureheight*0})
          {\includegraphics[width=\figurewidth]{./fig/cifar/relu_known_sample_high_uncertainty/image\x.png}};

      \foreach \x [count=\i] in {0,1,2,3,4,5,6}
        \node[draw=white,fill=black!20,minimum size=\figurewidth,inner sep=0pt]
          (\i) at ({\figurewidth*\i},{-\figureheight*1})
          {\includegraphics[width=\figurewidth]{./fig/cifar/relu_unknown_sample_low_uncertainty/image\x.png}};

      \foreach \x [count=\i] in {True: ship\\plane: 0.76,True: frog\\bird: 1.00,True: truck\\bird: 0.94,True: ship\\plane: 1.00,True: dog\\cat: 1.00,True: frog\\bird: 0.83,True: frog\\cat: 1.00}
        \node[align=center]
          (\i) at ({\figurewidth*\i},{-\figureheight*1.75})
          {\x};          
          
    \end{tikzpicture}}
    {Highest/lowest variance test samples (ReLU activation)}  
  \end{subfigure}
  \hfill
  \begin{subfigure}[t]{.49\textwidth}
    \centering
    \setlength{\figurewidth}{.18\textwidth}
    \setlength{\figureheight}{\figurewidth}
    \resizebox{\textwidth}{!}{
    \begin{tikzpicture}[inner sep=0]

      \tiny

      \node[rotate=90,text width=1.5\figureheight,align=center] at (.25,.35) {\tiny Highest uncertainty (known classes)};
      \node[rotate=90,text width=1.5\figureheight,align=center] at (.25,-1.65) {\tiny Lowest uncertainty (unknown classes)};

      \foreach \x [count=\i] in {True: cat\\cat: 0.82,True: cat\\cat: 0.84,True: cat\\cat: 0.64,True: bird\\cat: 0.66,True: cat\\cat: 0.74,True: cat\\cat: 0.75,True: car\\car: 0.56}
        \node[align=center]
          (\i) at ({\figurewidth*\i},{\figureheight*0.75})
          {\tiny \x};

      \foreach \x [count=\i] in {0,1,2,3,4,5,6}
        \node[draw=white,fill=black!20,minimum size=\figurewidth,inner sep=0pt]
          (\i) at ({\figurewidth*\i},{-\figureheight*0})
          {\includegraphics[width=\figurewidth]{./fig/cifar/matern52_known_sample_high_uncertainty/image\x.png}};

      \foreach \x [count=\i] in {0,1,2,3,4,5,6}
        \node[draw=white,fill=black!20,minimum size=\figurewidth,inner sep=0pt]
          (\i) at ({\figurewidth*\i},{-\figureheight*1})
          {\includegraphics[width=\figurewidth]{./fig/cifar/matern52_unknown_sample_low_uncertainty/image\x.png}};

      \foreach \x [count=\i] in {True: truck\\car: 0.98,True: ship\\plane: 0.98,True: frog\\bird: 0.98,True: truck\\car: 0.98,True: truck\\car: 0.98,True: dog\\cat: 0.98,True: truck\\plane: 0.98}
        \node[align=center]
          (\i) at ({\figurewidth*\i},{-\figureheight*1.75})
          {\tiny \x};          
          
    \end{tikzpicture}}
   {Highest/lowest variance test samples (Mat\'ern-5/2 activation)}    
  \end{subfigure}   
  \caption{OOD example on CIFAR-10 with 5 classes (`known') in training and all 10 in testing. Top:~Predictive variance histograms for known/unknown test class inputs, where the ReLU activation shows no separation, and the Mat\'ern shows gradually more separation with decreasing $\nu$. Bottom: Test samples from ends of the histograms (true and predicted label + class prob.). For the Mat\'ern, uncertain known and certain unknown samples feel intuitive (good calib.), while the results seem arbitrary for the ReLU. See \cref{app:cifar} for results for more activation functions.}  
  \label{fig:cifar}
  \vspace*{-1em}
\end{figure}

\paragraph{Out-of-distribution characterization with CIFAR-10}
As a rule of thumb the uncertainty of OOD samples should be high and uncertainty of in-distribution samples should be low. This corresponds to predicting confidently only for samples within the domain familiar to the model through training.
\cref{fig:cifar} shows the results for the experiments on the CIFAR-10 data set. Each model was trained with only images of five classes $\{\text{plane}, \text{car}, \text{bird}, \text{cat}, \text{deer}\}$. During testing, images from all 10 classes were present (now including also $\{\text{ship}, \text{truck}, \text{frog}, \text{dog}, \text{horse}\}$). In the histograms in \cref{fig:cifar}, samples from classes present during training are marked with green and samples from classes not used in training are marked with red. The histograms use the standard deviation of the DNN output across MC dropout samples as the measure of uncertainty.
\cref{fig:cifar} also shows examples of classified images. The top row shows images from known classes with the highest uncertainty for both models and the bottom row shows images from unknown classes with the lowest uncertainty. For each image, the correct label is shown, along with the label predicted by the DNN and its probability. Both models use the GoogLeNet~\cite{szegedy2015going} CNN architecture with one additional fully connected layer before the final classification layer. The difference between the tested models is only the activation function used in this additional layer. For both models, pre-trained weights are used except for the additional layer and the final classification layer. Both have an accuracy of 97\%. The Mat\'ern-5/2 model gives a mean NLPD of $0.11$ on the test set known classes (0.18 for the ReLU).

\begin{figure}[!t]
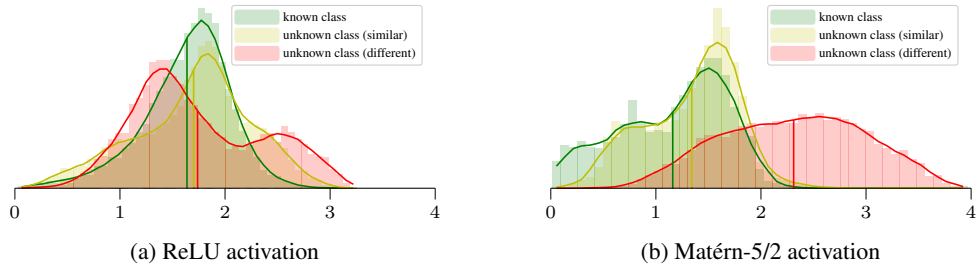

  \scriptsize
  \pgfplotsset{axis on top,scale only axis,width=\figurewidth,height=\figureheight}
  \pgfplotsset{legend style={inner xsep=1pt, inner ysep=1pt, row sep=0pt},legend style={rounded corners=1pt},legend style={nodes={scale=1.0, transform shape}}}  
  \setlength{\figurewidth}{.4\textwidth}
  \setlength{\figureheight}{.45\figurewidth}
  \begin{subfigure}[t]{.49\textwidth}
    \centering
    \input{./fig/MCdrop_relu.tex}\\[0em]
    \caption{ReLU activation}
  \end{subfigure}
  \hfill
  \begin{subfigure}[t]{.49\textwidth}
    \centering
    \input{./fig/MCdrop_custacti.tex}\\[0em]
    \caption{Mat\'ern-5/2 activation}
  \end{subfigure}
  \caption{Predictive entropy histograms for OOD radar emitter classification test samples, where the Mat\'ern shows better separation both for the `similar' and `different' emitter sets.}  
  \label{fig:emitter}
  \vspace*{-1em}
\end{figure}

\paragraph{Black-box radio emitter classification}
Radar emitter classification is a method for aircraft recognition in aviation. Classical methods for this problem resort to table lookup. The recent popularity and success of deep learning models have made them an appealing method to be applied to emitter classification. However, the lack of reliable uncertainty estimates for deep learning models risks making overconfident incorrect decisions with irreversible consequences. We train a CNN to classify simulated radar signals with a realistic radar library size training set with 100 different emitters. In the testing phase, the models receive samples from the same classes that were used in training and from additional unknown classes. These unknown classes are divided into two groups: classes that are {\em similar} to (and easily confused with) the classes used during training, and classes that are very distinct from the training data. \cref{fig:emitter} shows histograms of predictive entropy as the uncertainty measure for these groups of test data classes. The tested models differ only by the activation function used in the fully connected layer before the final classification layer, and have similar overall accuracy (59.3\% vs.\ 59.5\%). The Mat\'ern activation functions, however, help the model have more appropriate uncertainty estimates.

\section{Discussion and Conclusions}
\label{sec:discuss}
We established a link between neural network activation functions and the widely-known Mat\'ern family of kernels (covariance functions). Our derivation took a control theory perspective to random neural networks, which has roots in early AI, but has been overlooked in recent years. Our experiments showed wide applicability and good practical performance in Bayesian deep learning tasks.

In the experiments, inference was done by MC dropout. We point out that even if recent studies have shown MC dropout to have poor uncertainty estimation on OOD inputs \cite{Snoek+Ovadia19_trust,shafaei18_less_biased_evaluat_out_of,ritter2018a_kfac_laplace} or even in-distribution inputs \cite{Foong+Burt+Li:2019}, they used ReLU activation functions, where this behaviour is encoded into the model. In addition to MC dropout, Mat\'ern activations can be used in conjunction with more advanced inference methods (\eg, \cite{maddox19_SWAG}).
The selection of activation functions \cite{ramachandran2017searching} is typically on the basis of accuracy alone, while we suggest that the choice should also be considered from an uncertainty quantification angle. However, the Mat\'ern activation function is not monotonic, and thus the error surface associated with a single-layer model is not guaranteed to be convex \cite{wu2009global}. It also saturates to $0$ for most $\vx$, and could thus be difficult to optimize \cite{goodfellow2016deep}. Even so, usually choosing an appropriate learning rate was sufficient to achieve successfull training, and the only issues we encountered in the experiments occurred when using the (non-differentiable) exponential activation function. Both these properties are shared with RBF activations. We can, however, avoid some pathologies in RBFs. \citet{Stein:1999} argues that an infinite smoothness assumption is unrealistic for modelling many physical processes, and \citet{Duvenaud:cookbook} discusses several pathologies in infinitely smooth RBF and RQ models and points out that model misspecification is typically hard to spot. We thus consider the Mat\'ern activations to be a convenient and principled building block for encoding continuity, (non-infinite) smoothness, and stationarity assumptions in neural networks.

Example codes implementing the proposed methods in this paper are available at \url{https://github.com/AaltoML/stationary-activations}.

\section*{Broader Impact}
We propose a new building block to help quantify uncertainty in deep learning. This contributes to creating methods in artificial intelligence that know what they do not know, which is important in safety-critical applications and in creating more robust and reliable systems. Such safety-critical applications include, for example, automatic medical diagnosis, self-driving vehicles, and general tasks for decision-making under uncertainty \cite{amodei16_concrete}.

The contribution of this paper is in showing an explicit connection between two different paradigms in machine learning: we derive a non-linear activation function for neural networks which behaves as the widely used Mat\'ern class of Gaussian process priors. This  link can help build neural network models that are more robust and less vulnerable to out-of-distribution (OOD) data---which often occurs naturally in real-world settings \citep{ren19_likel_ratios_out_of_distr_detec,choi18_waic_but_why} or by potentially malicious construction, \eg, adversarial attacks \citep{szegedy14_intriguing_properties}---by making the neural networks behave more similarly to Gaussian process models, which have appealing properties in terms of well-calibrated uncertainty estimates and direct ways of including {\it a~priori} knowledge, but typically do not directly scale to all kinds of applications and large data sets. However, we only show theoretical guarantees of this link to hold in the limit of infinitely wide neural networks with one hidden layer. In finite-size models we can only demonstrate the benefits empirically.

This paper is concerned with foundational research which is expected to have an impact across application areas. The example applications in the paper underline the wide variety of possible applications ranging from simple classification tasks to out-of-distribution class detection in image and radar emitter classification.

\begin{ack}
We acknowledge the computational resources provided by the Aalto Science-IT project. This research was supported by the Academy of Finland grants 324345 and 308640, and Saab Finland~Oy. We thank William J.\ Wilkinson, Paul E.\ Chang, and the anonymous reviewers for feedback on the manuscript. Additionally, we thank Henrik Holter for helpful discussions.
\end{ack}

\phantomsection%
\addcontentsline{toc}{section}{References}
\begingroup
\small
\bibliographystyle{abbrvnat}
\bibliography{bibliography}

\begin{thebibliography}{72}
\providecommand{\natexlab}[1]{#1}
\providecommand{\url}[1]{\texttt{#1}}
\expandafter\ifx\csname urlstyle\endcsname\relax
  \providecommand{\doi}[1]{doi: #1}\else
  \providecommand{\doi}{doi: \begingroup \urlstyle{rm}\Url}\fi

\bibitem[Akhiezer and Glazman(1993)]{Akhiezer+Glazman:1993}
N.~I. Akhiezer and I.~M. Glazman.
\newblock \emph{Theory of Linear Operators in Hilbert Space}.
\newblock Dover, New York, 1993.

\bibitem[Amodei et~al.(2016)Amodei, Olah, Steinhardt, Christiano, Schulman, and
  Man{\'e}]{amodei16_concrete}
D.~Amodei, C.~Olah, J.~Steinhardt, P.~Christiano, J.~Schulman, and D.~Man{\'e}.
\newblock Concrete problems in {AI} safety.
\newblock \emph{arXiv preprint arXiv:1606.06565}, 2016.

\bibitem[Arora et~al.(2018)Arora, Basu, Mianjy, and
  Mukherjee]{arora2018understanding}
R.~Arora, A.~Basu, P.~Mianjy, and A.~Mukherjee.
\newblock Understanding deep neural networks with rectified linear units.
\newblock In \emph{International Conference on Learning Representations
  (ICLR)}, 2018.

\bibitem[Blundell et~al.(2015)Blundell, Cornebise, Kavukcuoglu, and
  Wierstra]{blundell2015weight}
C.~Blundell, J.~Cornebise, K.~Kavukcuoglu, and D.~Wierstra.
\newblock Weight uncertainty in neural networks.
\newblock In F.~Bach and D.~Blei, editors, \emph{Proceedings of the 32nd
  International Conference on International Conference on Machine Learning
  (ICML)}, volume~37 of \emph{Proceedings of Machine Learning Research}, pages
  1613--1622. PMLR, 2015.

\bibitem[Boult et~al.(2019)Boult, Dhamija, Gunther, Henrydoss, and
  Scheirer]{Boult19}
T.~Boult, A.~Dhamija, M.~Gunther, J.~Henrydoss, and W.~J. Scheirer.
\newblock Learning and the unknown: {S}urveying steps toward open world
  recognition.
\newblock In \emph{Proceedings of the AAAI Conference on Artificial
  Intelligence (AAAI)}, volume~33, pages 9801--9807, 2019.

\bibitem[{Bradshaw} et~al.(2017){Bradshaw}, {Matthews}, and
  {Ghahramani}]{bradshaw17_adver_examp_uncer_trans_testin}
J.~{Bradshaw}, A.~G. d.~G. {Matthews}, and Z.~{Ghahramani}.
\newblock Adversarial examples, uncertainty, and transfer testing robustness in
  {G}aussian process hybrid deep networks.
\newblock \emph{arXiv preprint arXiv:1707.02476}, 2017.

\bibitem[Broomhead and Lowe(1988)]{broomhead1988multivariable}
D.~S. Broomhead and D.~Lowe.
\newblock Multivariable functional interpolation and adaptive networks.
\newblock In \emph{Complex Systems}, volume~2, pages 321--355, 1988.

\bibitem[Cho and Saul(2009)]{cho+saul:2009}
Y.~Cho and L.~K. Saul.
\newblock Kernel methods for deep learning.
\newblock In \emph{Advances in Neural Information Processing Systems 22
  (NIPS)}, pages 342--350. Curran Associates, Inc., 2009.

\bibitem[{Choi} et~al.(2018){Choi}, {Jang}, and {Alemi}]{choi18_waic_but_why}
H.~{Choi}, E.~{Jang}, and A.~A. {Alemi}.
\newblock {WAIC}, but why? {G}enerative ensembles for robust anomaly detection.
\newblock \emph{arXiv preprint arXiv:1810.01392}, 2018.

\bibitem[Cortes and Vapnik(1995)]{cortes1995support}
C.~Cortes and V.~Vapnik.
\newblock Support-vector networks.
\newblock \emph{Machine Learning}, 20\penalty0 (3):\penalty0 273--297, 1995.

\bibitem[Cressie(1991)]{Cressie:1991}
N.~A. Cressie.
\newblock \emph{Statistics for Spatial Data}.
\newblock Wiley series in probability and mathematical statistics.
  Wiley-Interscience, 1991.

\bibitem[Croce and Hein(2018)]{croce2018randomized}
F.~Croce and M.~Hein.
\newblock A randomized gradient-free attack on {ReLU} networks.
\newblock In \emph{German Conference on Pattern Recognition}, pages 215--227.
  Springer, 2018.

\bibitem[Da~Prato and Zabczyk(1992)]{DaPrato:1992}
G.~Da~Prato and J.~Zabczyk.
\newblock \emph{Stochastic Equations in Infinite Dimensions}, volume~45 of
  \emph{Encyclopedia of Mathematics and its Applications}.
\newblock Cambridge University Press, 1992.

\bibitem[de~G.~Matthews et~al.(2018)de~G.~Matthews, Hron, Rowland, Turner, and
  Ghahramani]{matthews2018gaussian}
A.~G. de~G.~Matthews, J.~Hron, M.~Rowland, R.~E. Turner, and Z.~Ghahramani.
\newblock Gaussian process behaviour in wide deep neural networks.
\newblock In \emph{International Conference on Learning Representations
  (ICLR)}, 2018.

\bibitem[Dhamija et~al.(2018)Dhamija, G\"{u}nther, and
  Boult]{dhamija18_reduc_networ_agnos}
A.~R. Dhamija, M.~G\"{u}nther, and T.~Boult.
\newblock Reducing network agnostophobia.
\newblock In \emph{Advances in Neural Information Processing Systems 31
  (NeurIPS)}, pages 9157--9168. Curran Associates, Inc., 2018.

\bibitem[Duvenaud()]{Duvenaud:cookbook}
D.~Duvenaud.
\newblock The kernel cookbook: {A}dvice on covariance functions.
\newblock URL \url{https://www.cs.toronto.edu/~duvenaud/cookbook/}.

\bibitem[Duvenaud(2014)]{duvenaud2014automatic}
D.~Duvenaud.
\newblock \emph{Automatic Model Construction with {G}aussian Processes}.
\newblock PhD thesis, University of Cambridge, Cambridge, UK, 2014.

\bibitem[Fasshauer(2011)]{fasshauer:2011}
G.~E. Fasshauer.
\newblock Positive definite kernels: past, present and future.
\newblock \emph{Dolomites Research Notes on Approximation}, 4:\penalty0 21--63,
  2011.

\bibitem[Flam-Shepherd et~al.(2017)Flam-Shepherd, Requeima, and
  Duvenaud]{flam2017mapping}
D.~Flam-Shepherd, J.~Requeima, and D.~Duvenaud.
\newblock Mapping {G}aussian process priors to {B}ayesian neural networks.
\newblock In \emph{NIPS Workshop on Bayesian Deep Learning}, 2017.

\bibitem[Foong et~al.(2020)Foong, Burt, Li, and Turner]{Foong+Burt+Li:2019}
A.~Y. Foong, D.~R. Burt, Y.~Li, and R.~E. Turner.
\newblock On the expressiveness of approximate inference in {B}ayesian neural
  networks.
\newblock In \emph{Accepted for publication in Advances in Neural Information
  Processing Systems (NeurIPS)}, 2020.

\bibitem[Gal and Ghahramani(2016)]{Gal+Ghahramani:2016}
Y.~Gal and Z.~Ghahramani.
\newblock Dropout as a {B}ayesian approximation: {R}epresenting model
  uncertainty in deep learning.
\newblock In \emph{Proceedings of The 33rd International Conference on Machine
  Learning (ICML)}, volume~48 of \emph{Proceedings of Machine Learning
  Research}, pages 1050--1059. PMLR, 20--22 Jun 2016.

\bibitem[Gardner et~al.(2018)Gardner, Pleiss, Weinberger, Bindel, and
  Wilson]{GPyTorch}
J.~Gardner, G.~Pleiss, K.~Q. Weinberger, D.~Bindel, and A.~G. Wilson.
\newblock {GPyTorch}: {B}lackbox matrix-matrix {G}aussian process inference
  with {GPU} acceleration.
\newblock In \emph{Advances in Neural Information Processing Systems 31
  (NeurIPS)}, pages 7576--7586. Curran Associates, Inc., 2018.

\bibitem[Genton(2001)]{genton2001classes}
M.~G. Genton.
\newblock Classes of kernels for machine learning: {A} statistics perspective.
\newblock \emph{Journal of Machine Learning Research}, 2:\penalty0 299--312,
  2001.

\bibitem[Glad and Ljung(2000)]{Glad+Ljung:2000}
T.~Glad and L.~Ljung.
\newblock \emph{Control Theory: Multivariable and Nonlinear Methods}.
\newblock Taylor \& Francis, New York, 2000.

\bibitem[Goodfellow et~al.(2015)Goodfellow, Shlens, and
  Szegedy]{goodfellow15explaining}
I.~Goodfellow, J.~Shlens, and C.~Szegedy.
\newblock Explaining and harnessing adversarial examples.
\newblock In \emph{International Conference on Learning Representations
  (ICLR)}, 2015.

\bibitem[Goodfellow et~al.(2016)Goodfellow, Bengio, and
  Courville]{goodfellow2016deep}
I.~Goodfellow, Y.~Bengio, and A.~Courville.
\newblock \emph{Deep Learning}.
\newblock MIT Press, 2016.

\bibitem[Hafner et~al.(2019)Hafner, Tran, Lillicrap, Irpan, and
  Davidson]{hafner18_noise_contr_prior_funct_uncer}
D.~Hafner, D.~Tran, T.~P. Lillicrap, A.~Irpan, and J.~Davidson.
\newblock Noise contrastive priors for functional uncertainty.
\newblock In \emph{Proceedings of the Thirty-Fifth Conference on Uncertainty in
  Artificial Intelligence (UAI)}, page 332. {AUAI} Press, 2019.

\bibitem[Haykin(1999)]{Haykin:1999}
S.~Haykin.
\newblock \emph{Neural Networks: A Comprehensive Foundation}.
\newblock Prentice Hall, Upper Saddle River, NJ, USA, second edition, 1999.

\bibitem[Hein et~al.(2019)Hein, Andriushchenko, and Bitterwolf]{hein2019relu}
M.~Hein, M.~Andriushchenko, and J.~Bitterwolf.
\newblock Why {ReLU} networks yield high-confidence predictions far away from
  the training data and how to mitigate the problem.
\newblock In \emph{Proceedings of the IEEE Conference on Computer Vision and
  Pattern Recognition (CVPR)}, pages 41--50, 2019.

\bibitem[Hendrycks and Dietterich(2019)]{hendrycks2018benchmarking}
D.~Hendrycks and T.~Dietterich.
\newblock Benchmarking neural network robustness to common corruptions and
  perturbations.
\newblock In \emph{International Conference on Learning Representations
  (ICLR)}, 2019.

\bibitem[Hendrycks and Gimpel(2017)]{hendrycks17baseline}
D.~Hendrycks and K.~Gimpel.
\newblock A baseline for detecting misclassified and out-of-distribution
  examples in neural networks.
\newblock \emph{International Conference on Learning Representations (ICLR)},
  2017.

\bibitem[Hendrycks et~al.(2019)Hendrycks, Mazeika, and
  Dietterich]{hendrycks2018deep}
D.~Hendrycks, M.~Mazeika, and T.~Dietterich.
\newblock Deep anomaly detection with outlier exposure.
\newblock In \emph{International Conference on Learning Representations
  (ICLR)}, 2019.

\bibitem[Hensman et~al.(2015)Hensman, Matthews, and
  Ghahramani]{Hensman+Matthews+Ghahramani:2015}
J.~Hensman, A.~Matthews, and Z.~Ghahramani.
\newblock Scalable variational {G}aussian process classification.
\newblock In \emph{Proceedings of the Eighteenth International Conference on
  Artificial Intelligence and Statistics (AISTATS)}, volume~38 of
  \emph{Proceedings of Machine Learning Research}, pages 351--360. PMLR, 2015.

\bibitem[Hensman et~al.(2018)Hensman, Durrande, and
  Solin]{hensman2018variational}
J.~Hensman, N.~Durrande, and A.~Solin.
\newblock Variational {F}ourier features for {G}aussian processes.
\newblock \emph{Journal of Machine Learning Research}, 18:\penalty0 1--52,
  2018.

\bibitem[Hern{\'a}ndez-Lobato and Adams(2015)]{hernandez2015probabilistic}
J.~M. Hern{\'a}ndez-Lobato and R.~Adams.
\newblock Probabilistic backpropagation for scalable learning of {B}ayesian
  neural networks.
\newblock In \emph{Proceedings of the 32nd International Conference on Machine
  Learning (ICML)}, volume~37 of \emph{Proceedings of Machine Learning
  Research}, pages 1861--1869. PMLR, 2015.

\bibitem[Hofmann et~al.(2008)Hofmann, Sch{\"o}lkopf, and
  Smola]{hofmann2008kernel}
T.~Hofmann, B.~Sch{\"o}lkopf, and A.~J. Smola.
\newblock Kernel methods in machine learning.
\newblock \emph{The Annals of Statistics}, pages 1171--1220, 2008.

\bibitem[Lakshminarayanan et~al.(2017)Lakshminarayanan, Pritzel, and
  Blundell]{Lakshminarayanan17_deep_ensembles}
B.~Lakshminarayanan, A.~Pritzel, and C.~Blundell.
\newblock Simple and scalable predictive uncertainty estimation using deep
  ensembles.
\newblock In \emph{Advances in Neural Information Processing Systems 30
  (NIPS)}, pages 6402--6413. Curran Associates, Inc., 2017.

\bibitem[L{\'a}zaro-Gredilla et~al.(2010)L{\'a}zaro-Gredilla,
  Qui{\~n}onero-Candela, Rasmussen, and Figueiras-Vidal]{lazaro2010sparse}
M.~L{\'a}zaro-Gredilla, J.~Qui{\~n}onero-Candela, C.~E. Rasmussen, and A.~R.
  Figueiras-Vidal.
\newblock Sparse spectrum {G}aussian process regression.
\newblock \emph{Journal of Machine Learning Research}, 11:\penalty0 1865--1881,
  2010.

\bibitem[Lee et~al.(2018)Lee, Bahri, Novak, Schoenholz, Pennington, and
  Sohl-Dickstein]{lee2018deep}
J.~Lee, Y.~Bahri, R.~Novak, S.~S. Schoenholz, J.~Pennington, and
  J.~Sohl-Dickstein.
\newblock Deep neural networks as {G}aussian processes.
\newblock In \emph{International Conference on Learning Representations
  (ICLR)}, 2018.

\bibitem[Liang et~al.(2018)Liang, Li, and Srikant]{liang2018enhancing}
S.~Liang, Y.~Li, and R.~Srikant.
\newblock Enhancing the reliability of out-of-distribution image detection in
  neural networks.
\newblock In \emph{International Conference on Learning Representations
  (ICLR)}, 2018.

\bibitem[Maddox et~al.(2019)Maddox, Izmailov, Garipov, Vetrov, and
  Wilson]{maddox19_SWAG}
W.~J. Maddox, P.~Izmailov, T.~Garipov, D.~P. Vetrov, and A.~G. Wilson.
\newblock A simple baseline for {B}ayesian uncertainty in deep learning.
\newblock In \emph{Advances in Neural Information Processing Systems 32
  (NeurIPS)}, pages 13132--13143. Curran Associates, Inc., 2019.

\bibitem[Mat{\'e}rn(1960)]{Matern:1960}
B.~Mat{\'e}rn.
\newblock Spatial variation: {S}tochastic models and their applications to some
  problems in forest surveys and other sampling investigations.
\newblock \emph{Meddelanden fr{\aa}n statens skogsforskningsinstitut},
  49:\penalty0 1--144, 1960.

\bibitem[Matthews et~al.(2017)Matthews, {van der Wilk}, Nickson, Fujii,
  {Boukouvalas}, {Le{\'o}n-Villagr{\'a}}, Ghahramani, and Hensman]{GPflow:2017}
A.~G. d.~G. Matthews, M.~{van der Wilk}, T.~Nickson, K.~Fujii,
  A.~{Boukouvalas}, P.~{Le{\'o}n-Villagr{\'a}}, Z.~Ghahramani, and J.~Hensman.
\newblock {{GP}flow: A {G}aussian process library using {T}ensor{F}low}.
\newblock \emph{Journal of Machine Learning Research}, 18\penalty0
  (40):\penalty0 1--6, 2017.

\bibitem[Mercer(1909)]{mercer1909functions}
J.~Mercer.
\newblock Functions of positive and negative type and their connection with the
  theory of integral equations.
\newblock \emph{Philosophical Transactions of the Royal Society of London.
  Series A, Containing Papers of a Mathematical or Physical Character}, pages
  415--446, 1909.

\bibitem[Minh et~al.(2006)Minh, Niyogi, and Yao]{minh2006mercer}
H.~Q. Minh, P.~Niyogi, and Y.~Yao.
\newblock Mercer's theorem, feature maps, and smoothing.
\newblock In \emph{International Conference on Computational Learning Theory},
  pages 154--168. Springer, 2006.

\bibitem[Nalisnick et~al.(2019)Nalisnick, Matsukawa, Teh, Gorur, and
  Lakshminarayanan]{nalisnick2018do}
E.~Nalisnick, A.~Matsukawa, Y.~W. Teh, D.~Gorur, and B.~Lakshminarayanan.
\newblock Do deep generative models know what they don't know?
\newblock In \emph{International Conference on Learning Representations
  (ICLR)}, 2019.

\bibitem[Neal(1995)]{Neal:1995}
R.~M. Neal.
\newblock \emph{Bayesian Learning for Neural Networks}.
\newblock PhD thesis, University of Toronto, Toronto, Canada, 1995.

\bibitem[Nguyen et~al.(2015)Nguyen, Yosinski, and Clune]{nguyen2015deep}
A.~Nguyen, J.~Yosinski, and J.~Clune.
\newblock Deep neural networks are easily fooled: High confidence predictions
  for unrecognizable images.
\newblock In \emph{Proceedings of the IEEE Conference on Computer Vision and
  Pattern Recognition (CVPR)}, pages 427--436, 2015.

\bibitem[Pearce et~al.(2019)Pearce, Tsuchida, Zaki, Brintrup, and
  Neely]{pearce2019expressive}
T.~Pearce, R.~Tsuchida, M.~Zaki, A.~Brintrup, and A.~Neely.
\newblock Expressive priors in {B}ayesian neural networks: {K}ernel
  combinations and periodic functions.
\newblock In \emph{Proceedings of The 35th Conference on Uncertainty in
  Artificial Intelligence (UAI)}, page~25. AUAI Press, 2019.

\bibitem[Poggio and Girosi(1990)]{poggio1990networks}
T.~Poggio and F.~Girosi.
\newblock Networks for approximation and learning.
\newblock In \emph{Proceedings of the IEEE}, volume~78, pages 1481--1497. IEEE,
  1990.

\bibitem[Rahimi and Recht(2008)]{rahimi2008}
A.~Rahimi and B.~Recht.
\newblock Random features for large-scale kernel machines.
\newblock In \emph{Advances in Neural Information Processing Systems 20
  (NIPS)}, pages 1177--1184. Curran Associates, Inc., 2008.

\bibitem[Ramachandran et~al.(2018)Ramachandran, Zoph, and
  Le]{ramachandran2017searching}
P.~Ramachandran, B.~Zoph, and Q.~V. Le.
\newblock Searching for activation functions.
\newblock In \emph{International Conference on Learning Representations (ICLR)
  Workshops}, 2018.

\bibitem[Rasmussen and Williams(2006)]{Rasmussen+Williams:2006}
C.~E. Rasmussen and C.~K.~I. Williams.
\newblock \emph{Gaussian Processes for Machine Learning}.
\newblock MIT Press, 2006.

\bibitem[Ren et~al.(2019)Ren, Liu, Fertig, Snoek, Poplin, Depristo, Dillon, and
  Lakshminarayanan]{ren19_likel_ratios_out_of_distr_detec}
J.~Ren, P.~J. Liu, E.~Fertig, J.~Snoek, R.~Poplin, M.~Depristo, J.~Dillon, and
  B.~Lakshminarayanan.
\newblock Likelihood ratios for out-of-distribution detection.
\newblock In \emph{Advances in Neural Information Processing Systems 32
  (NeurIPS)}, pages 14707--14718. Curran Associates, Inc., 2019.

\bibitem[Ritter et~al.(2018)Ritter, Botev, and
  Barber]{ritter2018a_kfac_laplace}
H.~Ritter, A.~Botev, and D.~Barber.
\newblock A scalable {L}aplace approximation for neural networks.
\newblock In \emph{International Conference on Learning Representations
  (ICLR)}, 2018.

\bibitem[S\"arkk\"a et~al.(2013)S\"arkk\"a, Solin, and
  Hartikainen]{Sarkka+Solin+Hartikainen:2013}
S.~S\"arkk\"a, A.~Solin, and J.~Hartikainen.
\newblock Spatiotemporal learning via infinite-dimensional {B}ayesian filtering
  and smoothing.
\newblock \emph{IEEE Signal Processing Magazine}, 30\penalty0 (4):\penalty0
  51--61, 2013.

\bibitem[Shafaei et~al.(2019)Shafaei, Schmidt, and
  Little]{shafaei18_less_biased_evaluat_out_of}
A.~Shafaei, M.~Schmidt, and J.~J. Little.
\newblock A less biased evaluation of out-of-distribution sample detectors.
\newblock In \emph{30th British Machine Vision Conference 2019}. {BMVA} Press,
  2019.

\bibitem[Sitzmann et~al.(2020)Sitzmann, Martel, Bergman, Lindell, and
  Wetzstein]{sitzmann2020implicit}
V.~Sitzmann, J.~N. Martel, A.~W. Bergman, D.~B. Lindell, and G.~Wetzstein.
\newblock Implicit neural representations with periodic activation functions.
\newblock In \emph{Accepted for publication in Advances in Neural Information
  Processing Systems (NeurIPS)}, 2020.

\bibitem[Snoek et~al.(2019)Snoek, Ovadia, Fertig, Lakshminarayanan, Nowozin,
  Sculley, Dillon, Ren, and Nado]{Snoek+Ovadia19_trust}
J.~Snoek, Y.~Ovadia, E.~Fertig, B.~Lakshminarayanan, S.~Nowozin, D.~Sculley,
  J.~Dillon, J.~Ren, and Z.~Nado.
\newblock Can you trust your model's uncertainty? {E}valuating predictive
  uncertainty under dataset shift.
\newblock In \emph{Advances in Neural Information Processing Systems 32
  (NeurIPS)}, pages 13991--14002. Curran Associates, Inc., 2019.

\bibitem[Solin and S{\"a}rkk{\"a}(2020)]{solin2020hilbert}
A.~Solin and S.~S{\"a}rkk{\"a}.
\newblock Hilbert space methods for reduced-rank {G}aussian process regression.
\newblock \emph{Statistics and Computing}, 30\penalty0 (2):\penalty0 419--446,
  2020.

\bibitem[Stein(1999)]{Stein:1999}
M.~L. Stein.
\newblock \emph{Interpolation of Spatial Data}.
\newblock Springer Series in Statistics. Springer, New York, 1999.

\bibitem[Sun et~al.(2019)Sun, Zhang, Shi, and Grosse]{Sun+Zhang+Shi:2019}
S.~Sun, G.~Zhang, J.~Shi, and R.~Grosse.
\newblock Functional variational {B}ayesian neural networks.
\newblock In \emph{International Conference on Learning Representations
  (ICLR)}, 2019.

\bibitem[Szegedy et~al.(2014)Szegedy, Zaremba, Sutskever, Bruna, Erhan,
  Goodfellow, and Fergus]{szegedy14_intriguing_properties}
C.~Szegedy, W.~Zaremba, I.~Sutskever, J.~Bruna, D.~Erhan, I.~Goodfellow, and
  R.~Fergus.
\newblock Intriguing properties of neural networks.
\newblock In \emph{International Conference on Learning Representations
  (ICLR)}, 2014.

\bibitem[Szegedy et~al.(2015)Szegedy, Liu, Jia, Sermanet, Reed, Anguelov,
  Erhan, Vanhoucke, and Rabinovich]{szegedy2015going}
C.~Szegedy, W.~Liu, Y.~Jia, P.~Sermanet, S.~Reed, D.~Anguelov, D.~Erhan,
  V.~Vanhoucke, and A.~Rabinovich.
\newblock Going deeper with convolutions.
\newblock In \emph{Proceedings of the IEEE Conference on Computer Vision and
  Pattern Recognition (CVPR)}, pages 1--9, 2015.

\bibitem[Tsuchida et~al.(2018)Tsuchida, Roosta, and
  Gallagher]{tsuchida2018invariance}
R.~Tsuchida, F.~Roosta, and M.~Gallagher.
\newblock Invariance of weight distributions in rectified {MLP}s.
\newblock In \emph{Proceedings of the 35th International Conference on Machine
  Learning (ICML)}, volume~80 of \emph{Proceedings of Machine Learning
  Research}, pages 4995--5004. PMLR, 2018.

\bibitem[Wahba(1990)]{wahba1990spline}
G.~Wahba.
\newblock \emph{Spline Models for Observational Data}.
\newblock Siam, 1990.

\bibitem[Wenzel et~al.(2020)Wenzel, Roth, Veeling, {\'S}wi{\k{a}}tkowski, Tran,
  Mandt, Snoek, Salimans, Jenatton, and Nowozin]{Wenzel+Roth+Veeling:2020}
F.~Wenzel, K.~Roth, B.~S. Veeling, J.~{\'S}wi{\k{a}}tkowski, L.~Tran, S.~Mandt,
  J.~Snoek, T.~Salimans, R.~Jenatton, and S.~Nowozin.
\newblock How good is the {B}ayes posterior in deep neural networks really?
\newblock In \emph{Proceedings of the 37th International Conference on Machine
  Learning (ICML)}, volume 119 of \emph{Proceedings of Machine Learning
  Research}. PMLR, 2020.

\bibitem[Williams(1997)]{williams97computing}
C.~K.~I. Williams.
\newblock Computing with infinite networks.
\newblock In \emph{Advances in Neural Information Processing Systems 9 (NIPS)},
  pages 295--301. MIT Press, 1997.

\bibitem[Williams(1998)]{williams1998computation}
C.~K.~I. Williams.
\newblock Computation with infinite neural networks.
\newblock \emph{Neural Computation}, 10\penalty0 (5):\penalty0 1203--1216,
  1998.

\bibitem[Williams and Seeger(2000)]{Williams+Seeger:2000}
C.~K.~I. Williams and M.~Seeger.
\newblock The effect of the input density distribution on kernel-based
  classifiers.
\newblock In \emph{Proceedings of the 17th International Conference on Machine
  Learning (ICML)}. Morgan Kaufmann, 2000.

\bibitem[Wilson et~al.(2016)Wilson, Hu, Salakhutdinov, and
  Xing]{Wilson+Hu+Salakhutdinov:2016}
A.~G. Wilson, Z.~Hu, R.~R. Salakhutdinov, and E.~P. Xing.
\newblock Stochastic variational deep kernel learning.
\newblock In \emph{Advances in Neural Information Processing Systems 29
  (NIPS)}, pages 2586--2594. Curran Associates, Inc., 2016.

\bibitem[Wu(2009)]{wu2009global}
H.~Wu.
\newblock Global stability analysis of a general class of discontinuous neural
  networks with linear growth activation functions.
\newblock \emph{Information Sciences}, 179\penalty0 (19):\penalty0 3432--3441,
  2009.

\end{thebibliography}
\endgroup

\clearpage

\appendix

\def\toptitlebar{\hrule height4pt \vskip .25in \vskip -\parskip} 
\def\bottomtitlebar{\vskip .29in \vskip -\parskip \hrule height1pt \vskip .09in} 

\newcommand{\nipstitle}[1]{{%
    \phantomsection\hsize\textwidth\linewidth\hsize%
    \vskip 0.1in%
    \toptitlebar%
    \begin{minipage}{\textwidth}%
        \centering{\Large\bf #1\par}%
    \end{minipage}%
    \bottomtitlebar%
    \addcontentsline{toc}{section}{#1}%
}}

\clearpage

\setcounter{section}{0}

\nipstitle{
    {Supplementary Material:}\\ 
    Stationary Activations for Uncertainty \\ Calibration in Deep Learning}

\pagestyle{empty}

This supplementary document is organized as follows. \cref{app:derivations} includes further details and derivations for the Methods section in the main paper. \cref{app:experiments} includes details on the experiments, baseline methods, data sets, and additional tables and result plots. 

\section{Derivations}
\label{app:derivations}

\subsection{Spectral Factorization of the Mat\'ern Spectral Density}
\label{app:matern-transfer}
We consider the stationary (and isotropic) Mat\'ern covariance function (the difference here compared to the main paper is just defining $\vr = \vx-\vx' \in \mathbb{R}^d$, the parameterization is the same as in \cite{Rasmussen+Williams:2006}):
\begin{equation}\label{eq:matern-r}
  \kappa_\mathrm{Mat.}(\vr) = \frac{2^{1-\nu}}{\Gamma(\nu)}\left(\sqrt{2\nu}\,\frac{\|\vr\|}{\ell}\right)^\nu \mathrm{K}_\nu\!\left(\sqrt{2\nu}\,\frac{\|\vr\|}{\ell}\right), 
\end{equation}
where $\nu$ is a smoothness and $\ell$ a characteristic length-scale parameter, $\mathrm{K}_\nu(\cdot)$ the modified Bessel function, and $\Gamma(\cdot)$ the gamma function. This covariance function can be equally presented as a spectral density function, as discussed in the main paper (Wiener--Khinchin theorem). We use the angular frequency Fourier transform convention which simplifies keeping track of the scaling terms. From the Fourier-duality, the spectral density function of \cref{eq:matern-r} can be recovered by the Fourier transform:
\begin{equation}
  S_\textrm{Mat}(\vomega) = \int \kappa_\textrm{Mat}(\vr) \,
       \exp( -\imag \, \vomega\T \vr) \dd \vr.
\end{equation}
Solving the integral gives
\begin{equation}\label{eq:mat-specdens-d}
  S_\textrm{Mat}(\vomega) = \frac{2^d \pi^{d/2} \Gamma(\nu+d/2) \lambda^{2\nu}}{\Gamma(\nu)}  \left( \lambda^2 + \norm{\vomega}^2\right)^{-(\nu+d/2)},
\end{equation}
where $\lambda=\sqrt{2\nu}/\ell$. We can collect the constants into $q^2 = {2^d \pi^{d/2} \Gamma(\nu+d/2) \lambda^{2\nu}}/{\Gamma(\nu)}$. We are interested in the response of the system under white noise, and thus we relax to $d=1$. This is a recurring task in signal processing and control theory (see, \cite{Glad+Ljung:2000} for a brief but comprehensive overview on the topic). Following the rationale in the main paper, we look at the transfer function that can be defined through the spectral density as $S(\omega) = G(\omega)\,q^2\,G(\omega)^*$, where $[\cdot]^*$ is the complex-conjugate. Starting from \cref{eq:mat-specdens-d}, we can now do the spectral factorization by manipulating the expression (recall that $\imag^2 = -1$ and $a^2 - b^2 = (a-b)(a+b)$):
\begin{align}
  S(\omega) &= q^2\, \left( \lambda^2+\omega^2 \right)^{-(\nu+1/2)} \\
            &= q^2\, \left( \lambda^2-(\imag\,\omega)^2 \right)^{-(\nu+1/2)} \\
            &= { \left( \lambda + \imag\,\omega \right)^{-(\nu+1/2)}} \, q^2 \,
    \left( \lambda- \imag\,\omega \right)^{-(\nu+1/2)} \\
            &= G(\imag\,\omega)\,q^2\,G(-\imag\,\omega), \label{eq:specfact-app}
\end{align}
which is the form we use in the main paper.

\subsection{Recovering the Mat\'ern Activation}
\label{app:matern-activation}
Following \cref{eq:specfact-app}, we can collect the transfer function of the corresponding stable part (see discussion in the main paper) as
\begin{equation}\label{eq:transfer-app}
  G(\imag \, \omega) = \left( \lambda+ \imag\,\omega \right)^{-(\nu+1/2)}.
\end{equation}
The remaining (power) spectral density (formally that of the driving white noise process $w(t)$ that captures the scaling coefficients) is
$q^2 = {2 \pi^{1/2} \lambda^{2\nu}\Gamma(\nu+1/2)}/{\Gamma(\nu)}$.
Now taking the inverse Laplace transform of the transfer function of $G(\imag \, \omega)$ yields (this can be shown by considering the simplified expression $\Laplace[x^{\alpha-1}\,\exp(-a\,x)](s) = \Gamma(\alpha)\,(s+a)^{-\alpha}$, for $\alpha>0$)
\begin{equation}
  \Laplace^{-1}[G(s)](x) = \frac{1}{\Gamma(\nu+1/2)} x^{\nu-1/2} \exp(-\lambda\,x),
\end{equation}
and by using the properties of the Laplace transform for expanding it to the real line, we   recover the transfer (or {\em activation}) in the input space:
\begin{equation}\label{eq:activation-app}
  \sigma(x)  = \frac{q}{\Gamma(\nu+1/2)}\,\Theta(x)\,x^{\nu-1/2}\,\exp(-\lambda\,x),
\end{equation}
where $\Theta(\cdot)$ is the Heaviside step function.

\subsection{Recovering RBF Activations in the Limit of $\nu\to\infty$}
\label{app:limit-rbf}
From the empirical results in \cref{fig:cov-act} it is clear that the Mat\'ern activations of form \cref{eq:activation-app} approach RBF activations as $\nu \to \infty$. However, it is not entirely trivial from \cref{eq:activation-app} why this is the case. Thus we provide the following high-level proof.

Let $x \in \mathbb{R}_+$ and $\ell=1$, such that \cref{eq:activation-app} can be simplified to
\begin{equation}\label{eq:sigma-real-plus}
  \sigma(x) \propto x^{\nu-1/2}\,\exp(-\lambda\,x),
\end{equation}
where $\lambda = \sqrt{2\nu}$. By rewriting 
\begin{equation}\label{eq:sigma-real-plus}
  \sigma(x) \propto x^{\nu-1+1/2}\,\exp(-\sqrt{2\nu}\,x)
\end{equation}
we can directly collect the terms in the following form ($\alpha=\nu+1/2$ and $\beta = \sqrt{2\nu}$)
\begin{equation}\label{eq:sigma-real-plus-2}
  \sigma(x) \propto x^{\alpha-1}\,\exp(-\beta\,x),
\end{equation}
which we recognize to have the same form as the probability density function of the gamma distribution
\begin{equation}\label{eq:gamma}
  f(x \mid \alpha, \beta) = \frac{\beta^\alpha\,x^{\alpha-1}\, \exp(-\beta\,x)}{\Gamma(\alpha)}, \quad \text{for } x>0, \alpha,\beta>0,
\end{equation}
where the parameterization is in terms of shape ($\alpha$) and rate ($\beta$). Because $\alpha \gg \beta$ ($\nu+1/2 \gg \sqrt{2\nu}$) as $\alpha \to \infty$, we can leverage the known result that the gamma distribution tends to a Gaussian distribution as $\alpha \to \infty$ with mean $\alpha/\beta$ and variance $\alpha/\beta^2 \to 1/2$ and thus recovering the RBF activation in the limit
\begin{equation}
  \lim_{\nu \to \infty} \sigma(x) = C\,\exp(-(x-c)^2),
\end{equation}
where $c = (\nu+1/2)/\sqrt{2\nu}$ and $C$ is a positive constant. Formally this means that the RBF activation is pushed towards positive infinity (see \cref{fig:activation}).

\subsection{Remark on the Role of $p(\vw)$}
In the derivation for the RBF-NN kernel \citet{williams97computing} considers the prior $p(\vw)$ on the weights to be a delta distribution such that $\vw = 1$. This is a special case of considering $\vw$ to uniformly randomly get values in $\{-1,1\}$ (`binary white'), which coincides with the presentation in \citet{williams97computing} and \citet{Rasmussen+Williams:2006}, due to the RBF activation function being an even function.

The common assumption of assuming $p(\vw)$ to be Gaussian, is covered in \cref{fig:covfun} (light coloured lines), where the smoothness properties (around origin for each $\nu$) are well preserved, but the tail behaviour is different (also agreeing with that of the RBF activation).

\subsection{Closed-form Expressions for the RBF Kernel}
\label{app:closed_form_RBF}
In \cref{sec:green}, we went through an alternative view through functional analysis. This approach proved to be tricky for the Mat\'ern class, even if it has been used for showing properties of the RBF (squared-exponential) kernel in the past. For the RBF many derivations simplify and can be done in closed form. This also helps in sanity checks for the Mat\'ern class by comparing the limit behaviour to the RBF. Thus we provide the following set of identities, which can be tedious to work out, but can be useful in analysing these problems in the spirit of \cref{sec:green}.

For the RBF covariance function we can form the associated eigenbasis in closed form. We consider the re-parameterized RBF/squared-exponential kernel/covariance function of the form $\kappa(x,x') = \exp(-\alpha^2 |x - x'|^2)$, where $\alpha^2 = 1/2\ell^2$. And a weight function (input density function) $w(x)=\frac{\beta}{\sqrt{\pi}} \exp(-\beta^2 x^2)$. For these choices the eigenbasis of the covariance function (or the covariance operator) can be written as follows \cite{fasshauer:2011}:
\begin{align} 
  \gamma_j &= \beta \alpha^{2j} \bigg( \frac{\beta^2}{2}\bigg(1+\sqrt{1+\bigg(\frac{2\alpha}{\beta}\bigg)^2}\bigg)+\alpha^2\bigg)^{-(j+\frac{1}{2})}, \\
  \varphi_j(x) &= \frac{\sqrt[8]{1+(\frac{2\alpha}{\beta})^2}}{\sqrt{2^j j!}} \,\exp\bigg({-\left( \sqrt{1+(\frac{2\alpha}{\beta})^2} - 1\right)\frac{\beta^2 x^2}{2}}\bigg) \,\mathrm{H}_j\bigg( \sqrt[4]{1+\left(\frac{2\alpha}{\beta} \right)^2}\beta x \bigg).
\end{align} 
These eigenvalues and eigenfunctions are given in terms of physicists' Hermite polynomials $\mathrm{H}_j(\cdot)$. The obtained eigenfunctions are orthonormal with respect to
\begin{equation}
  \int \varphi_i(\vx)\,\varphi_j(\vx)\,w(\vx) \dd \vx = \delta_{ij}.
\end{equation}
For example, if we choose $\alpha = \sqrt{2}$ and $\beta = 1$, the first four eigenvalues and eigenfunctions are
\begin{align}
  \varphi_0(\vx) &= \sqrt[4]{3}\,e^{-x^2}, & 
  \varphi_1(\vx) &= \sqrt[4]{108}\,x\,e^{-x^2}, \nonumber \\
  \varphi_2(\vx) &= \sqrt[4]{\frac{3}{4}}\,(6x^2-1)\,e^{-x^2}, &
  \varphi_3(\vx) &= 3\sqrt[4]{3}\,(2x^3-x)\,e^{-x^2},
\end{align}
with associated eigenvalues 
$ \gamma_0 = \frac{1}{2}, 
  \gamma_1 = \frac{1}{4}, 
  \gamma_2 = \frac{1}{8},$ and $
  \gamma_3 = \frac{1}{16}$.

\citet{poggio1990networks} do the derivation through Green's function for radial basis function networks, which recovers the RBF as the corresponding activation function. The RBF basis functions have a natural role, which can be seen through {\em Mercer's theorem} \cite{mercer1909functions}. The theorem states that any positive-definite kernel can be represented as the inner product between a fixed set of features, evaluated at $\vx$ and $\vx'$:
\begin{equation}\label{eq:Mercer}
  \kappa(\vx,\vx') = \vh(\vx)\T\vh(\vx').
\end{equation}
The RBF kernel on the real line has a representation in terms of infinitely many radial-basis functions of the form $h(x) \propto \exp(-\frac{1}{4\ell^2}(x-c_i)^2)$, but any particular feature representation of a kernel is not necessarily unique (see, \cite{minh2006mercer} and \cite{duvenaud2014automatic} for a more detailed overview).

\begin{figure}[!t]
  \centering
  \scriptsize
  \pgfplotsset{scale only axis,width=\figurewidth,height=\figureheight,y tick label style={rotate=90}}
  \pgfplotsset{every axis/.append style={grid style={line width=0.2pt,dotted,gray}}}
  \setlength{\figurewidth}{.25\textwidth}
  \setlength{\figureheight}{\figurewidth}  
  \begin{subfigure}[t]{.33\textwidth}
    \raggedleft
%
%
\begin{tikzpicture}

\begin{axis}[%
axis on top,
xmin=-4.06349206349206,
xmax=4.06349206349206,
xtick={-4, -2,  0,  2,  4},
xlabel={Input, $x$},
xmajorgrids,
y dir=reverse,
ymin=-4.06349206349206,
ymax=4.06349206349206,
ytick={-4, -2,  0,  2,  4},
ylabel={Input, $x'$},
ymajorgrids,
axis background/.style={fill=white},
legend style={legend cell align=left,align=left,draw=white!15!black},
width=\figurewidth,
height=\figureheight
]
\addplot [forget plot] graphics [xmin=-4.06349206349206,xmax=4.06349206349206,ymin=-4.06349206349206,ymax=4.06349206349206] {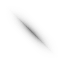};
\end{axis}
\end{tikzpicture}%
\\[0em]
    \caption{Mat\'ern-1/2}
  \end{subfigure}
  \hfill
  \begin{subfigure}[t]{.33\textwidth}
    \raggedleft
%
%
\begin{tikzpicture}

\begin{axis}[%
axis on top,
xmin=-4.06349206349206,
xmax=4.06349206349206,
xtick={-4, -2,  0,  2,  4},
xlabel={Input, $x$},
xmajorgrids,
y dir=reverse,
ymin=-4.06349206349206,
ymax=4.06349206349206,
ytick={-4,-2,0,2,4},
yticklabels={\empty},
ymajorgrids,
axis background/.style={fill=white},
legend style={legend cell align=left,align=left,draw=white!15!black},
width=\figurewidth,
height=\figureheight
]
\addplot [forget plot] graphics [xmin=-4.06349206349206,xmax=4.06349206349206,ymin=-4.06349206349206,ymax=4.06349206349206] {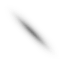};
\end{axis}
\end{tikzpicture}%
\\[0em]
    \caption{Mat\'ern-3/2}
  \end{subfigure}
  \hfill
  \begin{subfigure}[t]{.33\textwidth}
    \raggedleft
%
%
\begin{tikzpicture}

\begin{axis}[%
axis on top,
xmin=-4.06349206349206,
xmax=4.06349206349206,
xtick={-4, -2,  0,  2,  4},
xlabel={Input, $x$},
xmajorgrids,
y dir=reverse,
ymin=-4.06349206349206,
ymax=4.06349206349206,
ytick={-4,-2,0,2,4},
yticklabels={\empty},
ymajorgrids,
axis background/.style={fill=white},
legend style={legend cell align=left,align=left,draw=white!15!black},
width=\figurewidth,
height=\figureheight
]
\addplot [forget plot] graphics [xmin=-4.06349206349206,xmax=4.06349206349206,ymin=-4.06349206349206,ymax=4.06349206349206] {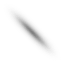};
\end{axis}
\end{tikzpicture}%
\\[0em]
    \caption{Mat\'ern-5/2}
  \end{subfigure}
  \caption{Gram matrices (colour map:~\protect\includegraphics[width=3em,height=0.6em]{fig/gray_r}) corresponding to different values for $\nu$ in the Mat\'ern-NN covariance function \cref{eq:Mat-NN} with particular choices for the hyperparameters ($\ell= 0.5$ and $\sigma^2_\mathrm{b} = 1^2$). The effect of the decay envelope is clearly visible when moving along the diagonal.}
\end{figure}

\section{Experiment Details}
\label{app:experiments}
The following sections provide further details on the experiment setup and implementation. Example codes are provided in separate files (see the accompanying \texttt{README} file) and they will also be available online at a later stage. 

\subsection{Illustrative Toy Examples}
To give an overview of the existing models and demonstrate how our model works, we provided two simple toy examples: a 2D classification and a 1D regression example. 

\paragraph{Classification}
In the classification example, we used the {\em Banana} data set (commonly even seen as a benchmarking data set). The Banana data set consists of 400 training samples, which are plotted as blue and orange circles in \cref{fig:teaser}. The training set has 183 samples from class 0 (blue color) and 217 samples from class 1 (orange colour). 

\begin{figure}[!t]
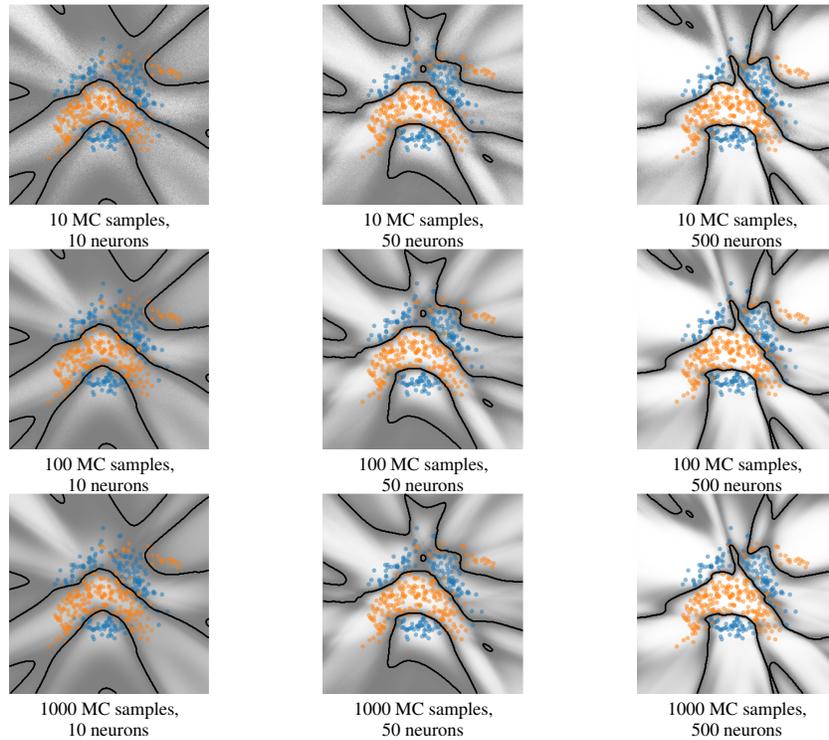

  \scriptsize
  \pgfplotsset{hide axis,scale only axis,width=\figurewidth,height=\figureheight}
  \setlength{\figurewidth}{.19\textwidth}
  \setlength{\figureheight}{\figurewidth}  
  \hspace*{\fill}
  \begin{subfigure}[t]{.195\textwidth}
    \centering
    \input{./fig/MLP_MCdrop_matern52_n10_mc10.tex}\\[0em]
    10 MC samples, \\ 10 neurons

  \end{subfigure}
  \hfill
  \begin{subfigure}[t]{.195\textwidth}
    \centering
    \input{./fig/MLP_MCdrop_matern52_n50_mc10.tex}\\[0em]
    10 MC samples, \\ 50 neurons

  \end{subfigure}
  \hfill
  \begin{subfigure}[t]{.195\textwidth}
    \centering
    \input{./fig/MLP_MCdrop_matern52_n500_mc10.tex}\\[0em]
    10 MC samples, \\ 500 neurons

  \end{subfigure} 
  \hspace*{\fill}
  \\
  \hspace*{\fill}
  \begin{subfigure}[t]{.195\textwidth}
    \centering
    \input{./fig/MLP_MCdrop_matern52_n10_mc100.tex}\\[0em]
    100 MC samples, \\ 10 neurons

  \end{subfigure}
  \hfill
  \begin{subfigure}[t]{.195\textwidth}
    \centering
    \input{./fig/MLP_MCdrop_matern52_n50_mc100.tex}\\[0em]
    100 MC samples, \\ 50 neurons

  \end{subfigure}
  \hfill
  \begin{subfigure}[t]{.195\textwidth}
    \centering
    \input{./fig/MLP_MCdrop_matern52_n500_mc100.tex}\\[0em]
    100 MC samples, \\ 500 neurons

  \end{subfigure} 
  \hspace*{\fill}
  \\
  \hspace*{\fill}
  \begin{subfigure}[t]{.195\textwidth}
    \centering
    \input{./fig/MLP_MCdrop_matern52_n10_mc1000.tex}\\[0em]
    1000 MC samples, \\10 neurons

  \end{subfigure}
  \hfill
  \begin{subfigure}[t]{.195\textwidth}
    \centering
    \input{./fig/MLP_MCdrop_matern52_n50_mc1000.tex}\\[0em]
    1000 MC samples, \\ 50 neurons

  \end{subfigure}
  \hfill
  \begin{subfigure}[t]{.195\textwidth}
    \centering
    \input{./fig/MLP_MCdrop_matern52_n500_mc1000.tex}\\[0em]
    1000 MC samples, \\ 500 neurons

  \end{subfigure}
  \hspace*{\fill}\\[-10pt]
  \caption{Illustrative comparisons on the {\em Banana} classification data set. The subfigures show decision boundaries and marginal predictive variance (low~\protect\includegraphics[width=3em,height=0.6em]{fig/gray_r}~high) for MLP neural network results (one hidden layer) with the network activation function being the Mat\'ern-$\frac{5}{2}$ activation. The rows show results with different numbers of MC dropout samples and the colums show results for different number of neurons in the hidden layer.}  
  \label{fig:banana_additional}
  \vspace*{-1em}
\end{figure}

We used a fully connected neural network architecture with one hidden layer of 50 nodes. This choice was partly to highlight the differences to the `infinitely wide' GP. For the same reason, we did not consider ensembling (the results were allowed to be noisy). The activation function of the hidden layer was set to represent the respective GP model kernel. For the Mat\'ern activation function we fixed the lengthscale $\ell=0.5$, which corresponds to scaling the inputs. The neural networks were trained for 2000 epochs with a batch size of 400 containing all training samples. The Adam optimizer was used with an initial learning rate of $0.02$, which was decayed by a factor 10 at epochs 250, 500, and 1000. Dropout with a rate of $0.2$ on the hidden layer was used during training to prevent overfitting and during testing to obtain MC dropout samples. For model testing a test sample grid of 300 by 300 samples in the range $(-3.75, 3.75)$ in both dimensions was created. To evaluate uncertainty during testing 5000 MC dropout samples were sampled for each grid sample (large number provides smoother figures). For each grid sample the MC dropout samples were averaged after applying the softmax function on the network output, and the resulting class probabilities $p$ and $q$ for the two classes were interpreted as a Bernoulli distribution standard deviation $\sigma = \sqrt{pq}$. This standard deviation was used as the uncertainty measure for the 2D classification task, plotted in \cref{fig:teaser}. Figure~\cref{fig:banana_additional} shows further tests on the {\em Banana} data set, visualizing the effect of changing the number of MC dropout samples and the number of neurons in the hidden layer of the network.

\paragraph{Regression}
For the 1D regression example, we generated a simple data set using the function $f(x) = \frac{1}{4}(x-\frac{1}{2})^3 + \frac{1}{2}x^2 - x + \frac{1}{5}$, which was chosen to generate differing function values at the two clusters and to have moderate curvature in the function at both clusters. The clusters were generated by sampling 100 values for $x$ at both clusters (200 samples in total) from Gaussian distributions with means of $-1$ and $1$, and with a standard deviation of $0.07$. The function $f(x)$ was evaluated at these values, and Gaussian noise with a standard deviation of $0.02$ was added. The resulting training samples for 1D regression are plotted as red dots in \cref{fig:1d}.
 
For simplicity, we used the same network architecture as in the classification example. Also the same dropout rate was applied during both training and testing. On the 1D regression task the network was trained for 2000 epochs with a batch size of 200 containing all training samples in a single batch. Adam optimizer with the same initial learning rate and learning rate decay schedule was used, as for 2D classification. The lengthscale was set to $\ell=1$ also for this regression example. To test the model an evenly spaced sample grid of 100 samples in the range $(-2.5, 2.5)$ was used. For each grid sample 1000 MC dropout samples were obtained to estimate uncertainty. The standard deviation of the network outputs accross MC dropout samples was used as the uncertainty measure (sigma in \cref{fig:1d}, plotted as light blue color). The training and evaluation process was repeated 20 times for each model and the average result is reported in \cref{fig:1d}.

\paragraph{GP baselines}
For the GP classification and regression baseline results we used GPflow~2 (\url{https://github.com/GPflow/GPflow}). The regression problem was a vanilla GP regression task, and for the classification task, we used the Bernoulli likelihood with VGP inference (corresponding to a non-sparse variant of \cite{Hensman+Matthews+Ghahramani:2015}). GPflow has the ArcCos kernel built-in, and we implemented the ERF-NN and RBF-NN kernels following the parametrization in \citet{Rasmussen+Williams:2006}. We used Adam for optimizing the hyperparameters and variational parameters. 

\begin{table}[t!]
  \caption{Further examples of UCI classification tasks, showing results for different choices for the covariance function (or correspondingly activation function). The setup in each sub table is the same and only the GP prior differs. Interestingly, the smoothness assumption of the prior seems to play a clear role, and the low-order Mat\'erns perform clearly better than the RBF.}
  \label{tbl:benchmarks-app}
  
  \begin{subfigure}{\textwidth}
  \caption{RBF: The neural network model outperforms the other methods except on the adult data set. However, overall performance using the RBF is worse for all models compared to when a lower-order Mat\'ern is used.}\label{tbl:benchmarks-app-a}
  \tiny
  \setlength{\tabcolsep}{0pt}
  \setlength{\tblw}{0.1\textwidth}  
  \begin{tabularx}{\textwidth}{l @{\extracolsep{\fill}} c @{\extracolsep{\fill}} c @{\extracolsep{\fill}} c @{\extracolsep{\fill}} C{\tblw} C{\tblw} | C{\tblw} C{\tblw} C{\tblw} C{\tblw} C{\tblw} C{\tblw}}
  \toprule
  \multicolumn{4}{l}{(10-fold cv)} & \multicolumn{2}{c}{SVGP} & \multicolumn{2}{c}{GPDNN} & \multicolumn{2}{c}{SV-DKL} &\multicolumn{2}{c}{RBF activation} \\
  & $n$ & $d$ & $c$ & NLPD & ACC & NLPD & ACC & NLPD & ACC & NLPD & ACC \\
  \midrule
  Adult & 45222 & 14 & 2 & $.341{\pm}.007$ & $.840{\pm}.006$ & $.431{\pm}.012$ & $.850{\pm}.005$ & $\bf.317{\pm}.006$ & $.854{\pm}.005$ & $.327{\pm}.012$ & $.853{\pm}.004$ \\
  Connect-4 & 67556 & 42 & 3 & $.611{\pm}.009$ & $.756{\pm}.005$ & $.782{\pm}.026$ & $.762{\pm}.008$ & $.501{\pm}.110$ & $.811{\pm}.049$ & $\bf.473{\pm}.010$ & $.817{\pm}.005$ \\  
  Covtype & 581912 & 54 & 7 & $.505{\pm}.004$ & $.782{\pm}.002$ & $.824{\pm}.063$ & $.816{\pm}.021$ & $.125{\pm}.073$ & $.952{\pm}.029$ & $\bf.116{\pm}.002$ & $.959{\pm}.001$ \\  
  Diabetes & 768 & 8 & 2 & $.509{\pm}.037$ & $.755{\pm}.057$ & $.615{\pm}.015$ & $.741{\pm}.041$ & $.695{\pm}.005$ & $.576{\pm}.086$ & $\bf.565{\pm}.052$ & $.718{\pm}.047$ \\ 
  \bottomrule
  \end{tabularx}
  \end{subfigure}
  
  \begin{subfigure}{\textwidth}
  \caption{Mat\'ern-5/2: The difference in the results given by this rather high-order Mat\'ern and the RBF are rather clear. Overall, the performance is similar as when a Mat\'ern-3/2 is used.}
  \tiny
  \setlength{\tabcolsep}{0pt}
  \setlength{\tblw}{0.1\textwidth}  
  \begin{tabularx}{\textwidth}{l @{\extracolsep{\fill}} c @{\extracolsep{\fill}} c @{\extracolsep{\fill}} c @{\extracolsep{\fill}} C{\tblw} C{\tblw} | C{\tblw} C{\tblw} C{\tblw} C{\tblw} C{\tblw} C{\tblw}}
  \toprule
  \multicolumn{4}{l}{(10-fold cv)} & \multicolumn{2}{c}{SVGP} & \multicolumn{2}{c}{GPDNN} & \multicolumn{2}{c}{SV-DKL} &\multicolumn{2}{c}{Mat\'ern activation} \\
  & $n$ & $d$ & $c$ & NLPD & ACC & NLPD & ACC & NLPD & ACC & NLPD & ACC \\
  \midrule
  Adult & 45222 & 14 & 2 & $.342{\pm}.007$ & $.841{\pm}.005$ & $.873{\pm}.116$ & $.804{\pm}.050$ & $\bf.315{\pm}.006$ & $.854{\pm}.005$ & $.317{\pm}.007$ & $.855{\pm}.004$ \\
  Connect-4 & 67556 & 42 & 3 & $.619{\pm}.010$ & $.754{\pm}.006$ & $1.74{\pm}.070$ & $.758{\pm}.009$ & $.462{\pm}.014$ & $.826{\pm}.006$ & $\bf.453{\pm}.009$ & $.827{\pm}.005$ \\  
  Covtype & 581912 & 54 & 7 & $.496{\pm}.003$ & $.786{\pm}.002$ & $.779{\pm}.032$ & $.822{\pm}.016$ & $\bf.101{\pm}.004$ & $.962{\pm}.002$ & $.115{\pm}.002$ & $.960{\pm}.001$ \\  
  Diabetes & 768 & 8 & 2 & $.507{\pm}.035$ & $.763{\pm}.055$ & $.608{\pm}.024$ & $.755{\pm}.042$ & $.693{\pm}.004$ & $.629{\pm}.073$ & $\bf .489{\pm}.074$ & $.768{\pm}.051$ \\
  \bottomrule
  \end{tabularx}
  \end{subfigure}
  
  \begin{subfigure}{\textwidth}
  \caption{Mat\'ern-3/2: Neural network model performs comparably or better compared to the GP or NN+GP hybrids.}
  \tiny
  \setlength{\tabcolsep}{0pt}
  \setlength{\tblw}{0.1\textwidth}  
  \begin{tabularx}{\textwidth}{l @{\extracolsep{\fill}} c @{\extracolsep{\fill}} c @{\extracolsep{\fill}} c @{\extracolsep{\fill}} C{\tblw} C{\tblw} | C{\tblw} C{\tblw} C{\tblw} C{\tblw} C{\tblw} C{\tblw}}
  \toprule
  \multicolumn{4}{l}{(10-fold cv)} & \multicolumn{2}{c}{SVGP} & \multicolumn{2}{c}{GPDNN} & \multicolumn{2}{c}{SV-DKL} &\multicolumn{2}{c}{Mat\'ern activation} \\
  & $n$ & $d$ & $c$ & NLPD & ACC & NLPD & ACC & NLPD & ACC & NLPD & ACC \\
  \midrule
  Adult & 45222 & 14 & 2 & $.344{\pm}.006$ & $.842{\pm}.005$ & $.435{\pm}.014$ & $.821{\pm}.037$ & $\bf.316{\pm}.006$ & $.855{\pm}.004$ & $\bf.316{\pm}.007$ & $.854{\pm}.005$ \\
  Connect-4 & 67556 & 42 & 3 & $.629{\pm}.010$ & $.750{\pm}.006$ & $.763{\pm}.018$ & $.768{\pm}.006$ & $.459{\pm}.016$ & $.827{\pm}.009$ & $\bf.450{\pm}.008$ & $.828{\pm}.004$ \\  
  Covtype & 581912 & 54 & 7 & $.494{\pm}.002$ & $.787{\pm}.002$ & $.722{\pm}.025$ & $.842{\pm}.008$ & $\bf.101{\pm}.005$ & $.962{\pm}.001$ & $.118{\pm}.003$ & $.958{\pm}.001$ \\  
  Diabetes & 768 & 8 & 2 & $.506{\pm}.034$ & $.759{\pm}.056$ & $.634{\pm}.012$ & $.744{\pm}.040$ & $.691{\pm}.005$ & $.507{\pm}.143$ & $\bf .486{\pm}.081$ & $.766{\pm}.044$ \\  
  \bottomrule
  \end{tabularx}
  \end{subfigure}
  
  \begin{subfigure}{\textwidth}
  \caption{Exponential: Very similar results to when other Mat\'ern options are used, except for our method, which suffers from the non-differentiability of the exponential activation function during training.}
  \tiny
  \setlength{\tabcolsep}{0pt}
  \setlength{\tblw}{0.1\textwidth}  
  \begin{tabularx}{\textwidth}{l @{\extracolsep{\fill}} c @{\extracolsep{\fill}} c @{\extracolsep{\fill}} c @{\extracolsep{\fill}} C{\tblw} C{\tblw} | C{\tblw} C{\tblw} C{\tblw} C{\tblw} C{\tblw} C{\tblw}}
  \toprule
  \multicolumn{4}{l}{(10-fold cv)} & \multicolumn{2}{c}{SVGP} & \multicolumn{2}{c}{GPDNN} & \multicolumn{2}{c}{SV-DKL} &\multicolumn{2}{c}{Exponential activation} \\
  & $n$ & $d$ & $c$ & NLPD & ACC & NLPD & ACC & NLPD & ACC & NLPD & ACC \\
  \midrule
  Adult & 45222 & 14 & 2 & $.349{\pm}.006$ & $.839{\pm}.005$ & $.428{\pm}.014$ & $.852{\pm}.006$ & $\bf.315{\pm}.005$ & $.855{\pm}.005$ & $.522{\pm}.084$ & $.783{\pm}.151$ \\
  Connect-4 & 67556 & 42 & 3 & $.680{\pm}.009$ & $.727{\pm}.005$ & $.747{\pm}.019$ & $.773{\pm}.006$ & $\bf.459{\pm}.015$ & $.827{\pm}.006$ & $.944{\pm}.056$ & $.670{\pm}.021$ \\  
  Covtype & 581912 & 54 & 7 & $.494{\pm}.002$ & $.791{\pm}.002$ & $.689{\pm}.128$ & $.811{\pm}.149$ & $\bf.101{\pm}.004$ & $.962{\pm}.001$ & $1.20{\pm}.009$ & $.491{\pm}.009$ \\  
  Diabetes & 768 & 8 & 2 & $.508{\pm}.030$ & $.762{\pm}.061$ & $\bf.611{\pm}.013$ & $.736{\pm}.048$ & $.691{\pm}.004$ & $.533{\pm}.137$ & $ 1.05{\pm}.307$ & $.443{\pm}.126$ \\
  \bottomrule
  \end{tabularx}
  \end{subfigure}
\end{table}

\begin{table}[t!]
  \caption{UCI classification tasks with the SIREN activation function \cite{sitzmann2020implicit} showing NLPD, accuracy and AUC metrics. The imposed prior model is different, so the results are not directly comparable with the other methods (\cref{tbl:benchmarks-app-a} being the closest).}
  \label{tbl:SIREN}
  \scriptsize
  \tiny
  \setlength{\tabcolsep}{0pt}
  \setlength{\tblw}{0.1\textwidth}  
  \begin{tabularx}{.5\textwidth}{l @{\extracolsep{\fill}} c @{\extracolsep{\fill}} c @{\extracolsep{\fill}} c @{\extracolsep{\fill}}  C{\tblw} C{\tblw} C{\tblw}}
  \toprule
  \multicolumn{4}{l}{(10-fold cv)} &\multicolumn{3}{c}{SIREN activation}\\
  & $n$ & $d$ & $c$  & NLPD & ACC & AUC \\
  \midrule
  Adult & 45222 & 14 & 2  & $.314{\pm}.006$ & $.854{\pm}.005$ & $.912{\pm}.004$ \\
  Connect-4 & 67556 & 42 & 3 & $.449{\pm}.008$ & $.825{\pm}.005$ & $.909{\pm}.003$\\  
  Covtype & 581912 & 54 & 7  & $.119{\pm}.002$ & $.957{\pm}.001$ & $.998{\pm}.000$\\  
  Diabetes & 768 & 8 & 2 & $.487{\pm}.066$ & $.771{\pm}.054$ & $.835{\pm}.054$\\  
  \bottomrule
  \end{tabularx}
  \vspace*{-1em}
\end{table}

\subsection{Benchmark Classification Tasks} 
\label{app:benchmark}
In the main paper, we considered four UCI benchmark classification tasks, where the aim was to compare our model to sophisticated previously published alternatives which aim to combine the flexibility of neural networks with GPs---in practice by taking neural network outputs as inputs to a GP model and performing joint learning and inference. We refer to these models as NN+GP hybrids. The rationale behind this experiment is that if our activations really emulate the behaviour of a GP with the corresponding covariance functions, we should be able to replace the hybrid GP part with a single layer with the corresponding Mat\'ern activation. In the experiments, we also used a standard GP classifier as baseline (SVGP).

For benchmark classification tasks we chose four UCI data sets: `adult', `connect-4', `covtype', and `diabetes'. The adult data set contains information of US citizens and the classification task is to predict whether the individual makes over or under \$50,000 in a year. The connect-4 data set contains positions in the game of connect-4 and the classification task is to predict whether player 1 will win or lose, or if the game will end in a draw. The covtype data set contains samples with cartographic variables describing 30 by 30 meter forest patches, and the classification task is to assign samples into one of seven possible forest cover type classes. The diabetes data set contains diagnostic data for patients and the classification task is to determine whether the patient has diabetes or not. The small diabetes data set ($n=768$) was chosen on purpose to see possible problems with small data.

The symbols $n$, $d$, and $c$ in \cref{tbl:benchmarks}, \cref{tbl:benchmarks-app}, and \cref{tbl:SIREN} represent the number of samples after removing missing values, the number of features in each sample, and the number of classes, respectively. 10-fold cross-validation was used to train and test each method. As preprocessing for all data sets categorical features were one-hot encoded and continuous features were normalized using standard scaling resulting in mean of 0 and variance of 1. For each fold of each method, the model was trained for 20 epochs with a batch size of 500 samples. For every method using a neural network as part of the model architecture, a fully connected network with layers d-1000-1000-500-50-c was used. The Adam optimizer was used as the model optimizer for all models except for SV-DKL, for which stochastic gradient descent optimizer was used with a weight decay of $0.0001$ and momentum of $0.9$. Training and testing was performed partially on CPUs and partially on GPUs (NVIDIA V100 and Tesla P100). On GPU the entire 10-fold cross-validation process for Mat\'ern activation functions takes roughly 5~seconds for the diabetes dataset, 1~minute for the adult and connect-4 datasets, and 13~minutes for the covtype dataset. Training the GP models is considerably slower.

\paragraph{SVGP}
(Stochastic Variational Gaussian Process) is a sparse GP method \cite{Hensman+Matthews+Ghahramani:2015}. We used a GPflow~2 \cite{GPflow:2017} implementation of SVGP to perform benchmark classification tasks. The learning rate was set to $0.01$. Bernoulli likelihood was used at the model outputs if the number of classes for the data set was two, otherwise a softmax likelihood was used. The number of inducing points was set to 500. This model acted as baseline and should not be directly compared to the NN+GP models (thus results separated by a vertical line in the table). 

\paragraph{GPDNN}
(Gaussian Process Deep Neural Networks, \cite{bradshaw17_adver_examp_uncer_trans_testin}) is a NN+GP hybrid model. We modified the GPflow reference implementation that was provided by Bradshaw (\url{https://gist.github.com/john-bradshaw/e6784db56f8ae2cf13bb51eec51e9057}). We used a RobustMax likelihood with $0.1$ epsilon on the model outputs. The number of inducing points and the number of active dimensions for the GP were set to 50. The learning rate was set to $0.001$. Training the GPDNN was surprisingly fragile and without an exactly tuned learning rate and the epsilon value, the training  diverged and NLPD values were very bad. The results in the tables are the best we could get for this model.

\paragraph{SV-DKL} (Stochastic Variational Deep Kernel Learning, \cite{Wilson+Hu+Salakhutdinov:2016}) is also a NN+GP hybrid model. The original SV-DKL publication considers a simple variant with pre-training and an additive GP. However, the SV-DKL framework has been improved as part of GPyTorch \cite{GPyTorch} to support general GP priors and end-to-end training. We used the reference implementation that is provided in the GPyTorch documentation (\url{https://gpytorch.readthedocs.io/en/latest/examples/06_PyTorch_NN_Integration_DKL/Deep_Kernel_Learning_DenseNet_CIFAR_Tutorial.html}). This implementation uses an approximate GP with inducing points placed on a grid. For the connect-4 data set a learning rate of $0.15$ was used, and for the other data sets the learning rate was set to $0.1$. Tuning the learning rates helped in achieving better convergence, but had only minor effects on the final numbers. The learning rate was used as  is for training the neural network parameters, and when training the GP model parameters the learning rate was scaled by a factor of $0.01$. At 10 and 15 epochs, the learning rate was decayed by a factor of 10. The number of inducing points and the number of active dimensions for the GP were set to 50. A softmax likelihood was used at the model outputs. During training the likelihood was sampled 8 times to obtain an approximate distribution and during evaluation likelihood was sampled 16 times. Training the SV-DKL was rather robust compared to the GPDNN.

\paragraph{Mat\'ern activation functions (ours)} were implemented using PyTorch. Similarly as in the simple examples, we fixed the scaling of the activations by choosing the lengthscale $\ell=0.5$. The learning rates were set to $0.00005$, $0.0002$, $0.0005$, and $0.0001$ for data sets adult, connect-4, covtype, and diabetes, respectively. Individual learning rates were set to achieve better convergence on each data set. At 10 and 15 epochs, the learning rate was decayed by a factor of 10. A softmax function was applied at the network output layer to obtain class probabilities. Dropout with a rate of $0.2$ was applied on the layer with 50 nodes during training to prevent overfitting and during testing to obtain MC dropout samples. For each test set sample 100 MC dropout samples were obtained to estimate uncertainty in the outputs. After applying the softmax function on the MC dropout sample outputs, these 100 samples were averaged to obtain single class probabilities for each test sample.

\paragraph{SIREN activation functions} \cite{sitzmann2020implicit} were implemented similar to the Mat\'ern activation functions using the same hyperparameters. The only difference to the Mat\'ern activation function implementation was the activation function used in the last hidden layer of the neural network.

\paragraph{Additional results} \cref{tbl:benchmarks-app} shows results on the benchmark classification tasks when different GP covariance functions and their corresponding activation functions are used. The table showing results for Mat\'ern-3/2 is identical to \cref{tbl:benchmarks} and shown here again for easier comparison. Looking at the results in this table we can observe that the neural network model using the activation function matching the GP kernel performs better or comparably to the GP and NN+GP hybrid models on all data sets, except when exponential kernel/activation is used. This is caused by the non-differentiability of the exponential activation which makes learning difficult for the neural network model. The GP and NN+GP hybrid models do not suffer from this problem as they are using the kernel instead of the activation function. Interestingly, the overall performance when an RBF is used is generally worse for all models when compared to most Mat\'ern alternatives, which can be expected as discussed in \cref{sec:discuss}. \cref{tbl:SIREN} shows results on the benchmark classification tasks when the SIREN activation is used.

To summarize, the results show that the Mat\'ern activation is capable of capturing the behaviour, accuracy, and uncertainty quantification of the NN+GP hybrid models, but without the slightly awkward stack of combining models. In the NN+GP hybrids, approximate inference and possible sparse methods have to be applied anyway.

\subsection{Out-of-distribution Characterization with CIFAR-10}
\label{app:cifar}
For image classification on the CIFAR-10 data set, we used the GoogLeNet~\cite{szegedy2015going} CNN architecture. We modified the standard architecture slightly by adding an additional linear layer of 512 nodes before the final classification layer. This makes the number of features in the end of the network be 1024-512-$c$, instead of 1024-$c$, where $c$ is the number of classes. We used pre-trained weights in every layer except in the additional linear layer and the final cltassification layer. Both the PyTorch base network architecture implementation and the pre-trained weights were based on those of \url{https://github.com/huyvnphan/PyTorch_CIFAR10}.

During training all pre-trained weights remained fixed. The difference between models compared on the CIFAR-10 task was only the choice of activation function in the additional linear layer of 512 nodes. Each model was trained for 100 epochs on the standard CIFAR-10 training set of 50,000 images but including only samples from five classes $\{\text{plane}, \text{car}, \text{bird}, \text{cat}, \text{deer}\}$ resulting in 25,000 training samples being used. A batch size of 128 and a default learning rate of $0.01$ with the Adam (with weight decay) optimizer were used for training. For all Mat\'ern activation functions a lengthscale $\ell=1$ was used. During testing, images from all 10 classes of the standard test set of 10,000 images were present (now including also $\{\text{ship}, \text{truck}, \text{frog}, \text{dog}, \text{horse}\}$). Dropout was applied on the additional linear layer with probability $0.2$ both to prevent overfitting during training and to implement MC dropout during testing. During testing 10 MC dropout samples were obtained for each test sample. Both training and testing were performed on GPUs (NVIDIA V100 and Tesla P100).

The test set of 10,000 images has 5000 images from `known' classes that were used during training and 5000 samples from `unknown' classes. To measure model performance we calculate classification accuracy including only samples from known classes, since classification accuracy on unknown classes is inevitably 0\% for all models. To further measure performance in terms of uncertainty estimates on known classes, we also calculate NLPD on test samples in known classes. To evaluate classification of test samples in unknown classes (where the standard NLPD measure is not applicable), we use a modified NLPD score:
\begin{equation}
  \text{For the `unknown' classes:}\qquad\mathrm{NLPD}_{\mathrm{mod}} = -\frac{1}{N}\sum_{i=1}^{N} \log\left(\frac{c}{c-1}(1-y_{\mathrm{max},i})\right),
\end{equation}
where $N$ is the number of samples, $c$ is the number of classes, and $y_{\mathrm{max},i}$ is the highest predicted class probability for sample $i$. This is similar to standard NLPD, differing only in how the predictive density is defined for unknown classes. The standard predicted class probability of the correct class can not be used, as the model can not predict any probability to unknown classes not specified for the model. Zero loss for a sample with this measure corresponds to predicting the maximum uncertainty prediction with class probabilities of $\frac{1}{c}$ for all classes, which is the best possible case for samples from unknown classes. This loss has infinite value for a sample, for which the model predicts some known class with 100\% confidence, despite the sample belonging to none of the known classes. This is not a standard metric, but provides interesting and intuitive numerical comparison for classifying samples from unknown classes.

The numerical results for using different activation functions in CIFAR-10 classification are shown in \cref{tbl:cifar}. For the results in the table, the network outputs were averaged across the 10 MC dropout samples and a softmax function was applied on the resulting mean network outputs to obtain class probabilities. Looking at the classification accuracy values, different models perform very similarly achieving very high accuracy except for when the exponential (Mat\'ern-1/2) or step activation is used. The model with the exponential activation manages to learn something meaningful with an accuracy of $37.6$\%, but the step activation has practically the accuracy of a random guess across the 5 known classes. This bad performance is explained by the non-differentiability of the Mat\'ern-1/2 (see \cref{fig:activation}) and step activations, which makes learning very challenging for the neural network.

In terms of NLPD on the known classes, the Mat\'ern class of activation functions have the best performance, but quite closely followed by the ERF and ReLU activations. For a practical example, a sample will have an NLPD of $0.1$ if it is predicted correctly with 90\% class probability. The NLPD values on the unknown class samples are strongly in favour of the RBF, Mat\'ern-5/2, and Mat\'ern-3/2 activations. 
The step function has the lowest NLPD on unknown class samples but this is arbitrary as the model using the step function has not learned anything meaningful based on classification accuracy and is just predicting all classes with almost equal probabilities, as can be seen in the predicted probabilities for the step activation in \cref{fig:cifar-app-2}.

After step activation, the second lowest NLPD value on unknown class samples is achieved by the Mat\'ern-5/2 activation at $0.754$. This value corresponds to predicting unknown samples with a maximum class probability of around 62\% on average. For comparison, the NLPD of $2.73$ for ERF corresponds to a maximum predicted class probability of 95\% on average for unknown samples. For ReLU activation the NLPD for unknown samples is infinite within numerical limits as the model predicts unknown class samples with 100\% confidence to belonging to some of the known classes.

The uncertainty histograms and examples of classified images in \cref{fig:cifar-app-2} show improvement in OOD sample uncertainty estimates when Mat\'ern activations are used. For the results in this figure, the uncertainty measure used was the standard deviation of MC dropout samples at the classification layer outputs before applying the softmax function. Ideally, we would like to be able to separate known and unknown class samples from each other based on their associated uncertainty. The histograms in \cref{fig:cifar-app-2} show that Mat\'ern class activation functions show good separation between samples of unknown and known classes. The ERF activation function also shows decent separation. The ReLU doesn't show practically any separation in uncertainty of the unknown and known classes. For the exponential and step activations, the histograms are not very meaningful as with these activations the models didn't manage to learn to properly classify the sample images. Looking at the samples of classified images for ReLU, all known class samples with highest uncertainty have a predicted class probability of $1.00$ and many of the unknown class samples with lowest uncertainty have quite low predicted class probabilities. This suggests that for ReLU, standard deviation of MC outputs and the predicted class probability do not correlate very well with each other as uncertainty measures.

\begin{table}[t!]
  \caption{CIFAR10 classification accuracy (on the `known' classes that were used during training) and mean NLPD values reported separately for the `known' classes and `unknown' OOD classes. For accuracy, larger is better, and NLPD smaller is better. For unknown classes the NLPD is calculated as described in \cref{app:cifar}.}
  \label{tbl:cifar}
  \scriptsize

  \setlength{\tabcolsep}{0pt}
  \setlength{\tblw}{0.1\textwidth}  
  \begin{tabularx}{\textwidth}{l @{\extracolsep{\fill}} C{\tblw} C{\tblw} C{\tblw} C{\tblw} C{\tblw} C{\tblw} C{\tblw} C{\tblw}}
  \toprule
  & RBF &Mat\'ern-5/2 & Mat\'ern-3/2 & Exponential & ERF & Step & ReLU & SIREN\\
  \midrule
  Accuracy (known class) & 0.973 & 0.974 & 0.973 & 0.376 & \bf0.975 & 0.217 & 0.973 & 0.972\\
  NLPD (known class) & \bf0.099 & 0.110 & 0.103 & 1.26 & 0.147 & 1.52 & 0.181 & 0.106\\
  NLPD (unknown class) & 0.801 & 0.754 & 0.896 & 1.22 & 2.73 & \bf0.114 & (${\sim}$inf) & 1.62\\   
  \bottomrule
  \end{tabularx}
\end{table}

\subsection{Black-box Radio Emitter Classification}
In the final example, we consider an application outside the standard benchmarking tasks, which underlines both the importance of uncertainty quantification and OOD characterization.
The data set for the radio emitter classification task was generated through simulations, as large radar data sets are not publicly available. However, the properties are well understood and the observations can be simulated. 

\paragraph{Data}
We use a simulator provided by Saab for creating a realistic data set. The simulation setup was tuned in collaboration with experts at Saab. Each simulated radar emitter sample has a carrier frequency, pulse width, and a series of 25 to 250 pulse arrival times. The pulse arrival times follow some pattern characteristic to the specific radar emitter. To set up a classification task we specify 100 simulated emitters with characteristic carrier frequencies, pulse widths, and pulse arrival time patterns. These characteristic parameters are partly overlapping between different simulated emitters to make the classification task harder. Each simulated emitter can also operate in 10 different modes of operation, each of which have slightly different characteristic parameters. For each mode of each emitter, we simulated 5 samples on 11 different noise levels. In total, this adds up to 55,000 training samples. The noise is added to the samples by disturbing the series of pulse arrival times, by either dropping pulses or by adding spurious pulses to the sequence.

In addition to the training set, we simulated a test set with three groups of samples. The first test sample group of 10,000 samples was generated from the same simulated emitters as the training set. The second test sample group of 10,000 samples was generated from simulated emitters that are unknown to the model, but that have their characteristic parameters resembling some of the simulated emitters in the training set, making these test samples hard to distinguish from familiar training samples. The third test sample group of 10,000 samples was generated from simulated emitters that are unknown to the model, and that have very different characteristic parameters to the simulated emitters in the training set, making these test samples easy to distinguish from samples that are from the training set emitters. Together these three test sample groups help to set up an OOD classification test, for which an ideal model can separate the samples in the two groups generated from unknown emitters, from the samples generated by the familiar training emitters. In this task separating the unknown emitter samples that more closely resemble the training set emitter samples should be harder than separating the samples that are from emitters with very different characteristics.

\paragraph{Model}
The neural network architecture used for this task is a CNN architecture with skip connections, combined with a small parallel fully connected network. The CNN part of the network is used to process the series of pulse arrival times with 1D convolutional layers, and the fully connected part is used to process the carrier frequency and pulse width information. The outputs of these two parallel networks are concatenated and a fully connected layer structure of 148-120-$c$ is used to provide the final classification ($c=100$). Dropout with a rate of $0.2$ is applied on the fully connected layer with 120 nodes. The difference between the two tested models is also the activation function used in this fully connected layer: either a ReLU activation or Mat\'ern-5/2 activation with a lengthscale $\ell=1$ was used. Dropout is active also during testing to provide MC dropout samples (100 MC dropout samples are obtained). The Adam optimizer with a learning rate of $0.01$ was used in training. The models are trained for 7 epochs with a batch size of 50 samples.

\begin{figure}[!p]
  \scriptsize
  \pgfplotsset{axis on top,scale only axis,width=\figurewidth,height=\figureheight}
  \pgfplotsset{legend style={inner xsep=1pt, inner ysep=1pt, row sep=0pt},legend style={rounded corners=1pt},legend style={nodes={scale=1.0, transform shape}}}
  \setlength{\figurewidth}{.35\textwidth}
  \setlength{\figureheight}{.45\figurewidth}
  \begin{subfigure}{\textwidth}
  \begin{subfigure}[t]{.49\textwidth}
    \centering
    \input{./fig/cifar/cifar10_rbf_app.tex}\\[0em]
  \end{subfigure}
  \hfill
    \begin{subfigure}[t]{.49\textwidth}
    \centering
    \setlength{\figurewidth}{.18\textwidth}
    \setlength{\figureheight}{\figurewidth}
    \resizebox{\textwidth}{!}{
    \begin{tikzpicture}[inner sep=0]

      \tiny

      \node[rotate=90,text width=1.5\figureheight,align=center] at (.25,.35) {\tiny Highest uncertainty (known classes)};
      \node[rotate=90,text width=1.5\figureheight,align=center] at (.25,-1.65) {\tiny Lowest uncertainty (unknown classes)};

      \foreach \x [count=\i] in {True: bird\\bird: 0.86,True: cat\\cat: 0.73,True: cat\\cat: 0.81,True: car\\car: 0.94,True: cat\\cat: 0.60,True: car\\car: 0.70,True: cat\\cat: 0.71}
        \node[align=center]
          (\i) at ({\figurewidth*\i},{\figureheight*0.75})
          {\x};

      \foreach \x [count=\i] in {0,1,2,3,4,5,6}
        \node[draw=white,fill=black!20,minimum size=\figurewidth,inner sep=0pt]
          (\i) at ({\figurewidth*\i},{-\figureheight*0})
          {\includegraphics[width=\figurewidth]{./fig/cifar/rbf_known_sample_high_uncertainty/image\x.png}};

      \foreach \x [count=\i] in {0,1,2,3,4,5,6}
        \node[draw=white,fill=black!20,minimum size=\figurewidth,inner sep=0pt]
          (\i) at ({\figurewidth*\i},{-\figureheight*1})
          {\includegraphics[width=\figurewidth]{./fig/cifar/rbf_unknown_sample_low_uncertainty/image\x.png}};

      \foreach \x [count=\i] in {True: dog\\cat: 0.98,True: truck\\car: 0.99,True: dog\\cat: 0.98,True: dog\\cat: 0.98,True: dog\\cat: 0.98,True: truck\\plane: 0.99,True: frog\\cat: 0.98}
        \node[align=center]
          (\i) at ({\figurewidth*\i},{-\figureheight*1.75})
          {\x};          
          
    \end{tikzpicture}}
    {Highest/lowest variance test samples (RBF activation)}  
  \end{subfigure}
  \caption{Results with RBF activation}
  \end{subfigure}
  \\[4em]
  \begin{subfigure}{\textwidth}
  \begin{subfigure}[t]{.49\textwidth}
    \centering
    \input{./fig/cifar/cifar10_matern52.tex}\\[0em]
  \end{subfigure}
  \hfill
  \begin{subfigure}[t]{.49\textwidth}
    \centering
    \setlength{\figurewidth}{.18\textwidth}
    \setlength{\figureheight}{\figurewidth}
    \resizebox{\textwidth}{!}{
    \begin{tikzpicture}[inner sep=0]

      \tiny

      \node[rotate=90,text width=1.5\figureheight,align=center] at (.25,.35) {\tiny Highest uncertainty (known classes)};
      \node[rotate=90,text width=1.5\figureheight,align=center] at (.25,-1.65) {\tiny Lowest uncertainty (unknown classes)};

      \foreach \x [count=\i] in {True: cat\\cat: 0.82,True: cat\\cat: 0.84,True: cat\\cat: 0.64,True: bird\\cat: 0.66,True: cat\\cat: 0.74,True: cat\\cat: 0.75,True: car\\car: 0.56}
        \node[align=center]
          (\i) at ({\figurewidth*\i},{\figureheight*0.75})
          {\tiny \x};

      \foreach \x [count=\i] in {0,1,2,3,4,5,6}
        \node[draw=white,fill=black!20,minimum size=\figurewidth,inner sep=0pt]
          (\i) at ({\figurewidth*\i},{-\figureheight*0})
          {\includegraphics[width=\figurewidth]{./fig/cifar/matern52_known_sample_high_uncertainty/image\x.png}};

      \foreach \x [count=\i] in {0,1,2,3,4,5,6}
        \node[draw=white,fill=black!20,minimum size=\figurewidth,inner sep=0pt]
          (\i) at ({\figurewidth*\i},{-\figureheight*1})
          {\includegraphics[width=\figurewidth]{./fig/cifar/matern52_unknown_sample_low_uncertainty/image\x.png}};

      \foreach \x [count=\i] in {True: truck\\car: 0.98,True: ship\\plane: 0.98,True: frog\\bird: 0.98,True: truck\\car: 0.98,True: truck\\car: 0.98,True: dog\\cat: 0.98,True: truck\\plane: 0.98}
        \node[align=center]
          (\i) at ({\figurewidth*\i},{-\figureheight*1.75})
          {\tiny \x};          
          
    \end{tikzpicture}}
   {Highest/lowest variance test samples (Mat\'ern-5/2 activation)}    
  \end{subfigure}
  \caption{Results with Mat\'ern-5/2 activation}
  \end{subfigure}
  \\[4em]
  \begin{subfigure}{\textwidth}
  \begin{subfigure}[t]{.49\textwidth}
    \centering
    \input{./fig/cifar/cifar10_matern32_app.tex}\\[0em]
  \end{subfigure}
  \hfill
    \begin{subfigure}[t]{.49\textwidth}
    \centering
    \setlength{\figurewidth}{.18\textwidth}
    \setlength{\figureheight}{\figurewidth}
    \resizebox{\textwidth}{!}{
    \begin{tikzpicture}[inner sep=0]

      \tiny

      \node[rotate=90,text width=1.5\figureheight,align=center] at (.25,.35) {\tiny Highest uncertainty (known classes)};
      \node[rotate=90,text width=1.5\figureheight,align=center] at (.25,-1.65) {\tiny Lowest uncertainty (unknown classes)};

      \foreach \x [count=\i] in {True: bird\\bird: 0.84,True: bird\\bird: 0.49,True: bird\\bird: 0.55,True: cat\\cat: 0.63,True: bird\\bird: 0.45,True: bird\\deer: 0.52,True: plane\\car: 0.71}
        \node[align=center]
          (\i) at ({\figurewidth*\i},{\figureheight*0.75})
          {\tiny \x};

      \foreach \x [count=\i] in {0,1,2,3,4,5,6}
        \node[draw=white,fill=black!20,minimum size=\figurewidth,inner sep=0pt]
          (\i) at ({\figurewidth*\i},{-\figureheight*0})
          {\includegraphics[width=\figurewidth]{./fig/cifar/matern32_known_sample_high_uncertainty/image\x.png}};

      \foreach \x [count=\i] in {0,1,2,3,4,5,6}
        \node[draw=white,fill=black!20,minimum size=\figurewidth,inner sep=0pt]
          (\i) at ({\figurewidth*\i},{-\figureheight*1})
          {\includegraphics[width=\figurewidth]{./fig/cifar/matern32_unknown_sample_low_uncertainty/image\x.png}};

      \foreach \x [count=\i] in {True: truck\\plane: 0.99,True: truck\\car: 0.99,True: dog\\cat: 0.99,True: dog\\cat: 0.99,True: truck\\car: 0.99,True: horse\\deer: 0.98,True: dog\\cat: 0.98}
        \node[align=center]
          (\i) at ({\figurewidth*\i},{-\figureheight*1.75})
          {\tiny \x};          
          
    \end{tikzpicture}}
   {Highest/lowest variance test samples (Mat\'ern-3/2 activation)}    
  \end{subfigure}
  \caption{Results with Mat\'ern-3/2 activation}
  \end{subfigure}
  \\[4em]
  \begin{subfigure}{\textwidth}
  \begin{subfigure}[t]{.49\textwidth}
    \centering
    \input{./fig/cifar/cifar10_matern12.tex}\\[0em]
  \end{subfigure}
  \hfill
  \begin{subfigure}[t]{.49\textwidth}
    \centering
    \setlength{\figurewidth}{.18\textwidth}
    \setlength{\figureheight}{\figurewidth}
    \resizebox{\textwidth}{!}{
    \begin{tikzpicture}[inner sep=0]

      \tiny

      \node[rotate=90,text width=1.5\figureheight,align=center] at (.25,.35) {\tiny Highest uncertainty (known classes)};
      \node[rotate=90,text width=1.5\figureheight,align=center] at (.25,-1.65) {\tiny Lowest uncertainty (unknown classes)};

      \foreach \x [count=\i] in {True: deer\\deer: 0.79,True: deer\\deer: 0.65,True: deer\\deer: 0.62 ,True: deer\\deer: 0.68,True: deer\\deer: 0.86,True: deer\\deer: 0.76,True: deer\\deer: 0.73}
        \node[align=center]
          (\i) at ({\figurewidth*\i},{\figureheight*0.75})
          {\x};

      \foreach \x [count=\i] in {0,1,2,3,4,5,6}
        \node[draw=white,fill=black!20,minimum size=\figurewidth,inner sep=0pt]
          (\i) at ({\figurewidth*\i},{-\figureheight*0})
          {\includegraphics[width=\figurewidth]{./fig/cifar/matern12_known_sample_high_uncertainty/image\x.png}};

      \foreach \x [count=\i] in {0,1,2,3,4,5,6}
        \node[draw=white,fill=black!20,minimum size=\figurewidth,inner sep=0pt]
          (\i) at ({\figurewidth*\i},{-\figureheight*1})
          {\includegraphics[width=\figurewidth]{./fig/cifar/matern12_unknown_sample_low_uncertainty/image\x.png}};

      \foreach \x [count=\i] in {True: horse\\car: 0.27,True: horse\\deer: 0.27,True: dog\\deer: 0.30,True: horse\\deer: 0.28,True: ship\\plane: 0.41,True: horse\\deer: 0.33,True: truck\\car: 0.69}
        \node[align=center]
          (\i) at ({\figurewidth*\i},{-\figureheight*1.75})
          {\x};          
          
    \end{tikzpicture}}
    {Highest/lowest variance test samples (Exponential activation)}  
  \end{subfigure}
  \caption{Results with the exponential (Mat\'ern-1/2) activation}
  \end{subfigure}
  \\[1em]

\end{figure}  
\begin{figure}[!p]\ContinuedFloat
  \scriptsize
  \pgfplotsset{axis on top,scale only axis,width=\figurewidth,height=\figureheight}
  \pgfplotsset{legend style={inner xsep=1pt, inner ysep=1pt, row sep=0pt},legend style={rounded corners=1pt},legend style={nodes={scale=1.0, transform shape}}}
  \setlength{\figurewidth}{.35\textwidth}
  \setlength{\figureheight}{.45\figurewidth}  
  \begin{subfigure}{\textwidth}
  \begin{subfigure}[t]{.49\textwidth}
    \centering
    \input{./fig/cifar/cifar10_erf.tex}\\[0em]
  \end{subfigure}
  \hfill
  \begin{subfigure}[t]{.49\textwidth}
    \centering
    \setlength{\figurewidth}{.18\textwidth}
    \setlength{\figureheight}{\figurewidth}
    \resizebox{\textwidth}{!}{
    \begin{tikzpicture}[inner sep=0]

      \tiny

      \node[rotate=90,text width=1.5\figureheight,align=center] at (.25,.35) {\tiny Highest uncertainty (known classes)};
      \node[rotate=90,text width=1.5\figureheight,align=center] at (.25,-1.65) {\tiny Lowest uncertainty (unknown classes)};

      \foreach \x [count=\i] in {True: cat\\deer: 0.71,True: cat\\cat: 0.76,True: bird\\bird: 0.64,True: cat\\cat: 0.93,True: cat\\bird: 0.78,True: car\\car: 0.67,True: deer\\deer: 0.86}
        \node[align=center]
          (\i) at ({\figurewidth*\i},{\figureheight*0.75})
          {\tiny \x};

      \foreach \x [count=\i] in {0,1,2,3,4,5,6}
        \node[draw=white,fill=black!20,minimum size=\figurewidth,inner sep=0pt]
          (\i) at ({\figurewidth*\i},{-\figureheight*0})
          {\includegraphics[width=\figurewidth]{./fig/cifar/erf_known_sample_high_uncertainty/image\x.png}};

      \foreach \x [count=\i] in {0,1,2,3,4,5,6}
        \node[draw=white,fill=black!20,minimum size=\figurewidth,inner sep=0pt]
          (\i) at ({\figurewidth*\i},{-\figureheight*1})
          {\includegraphics[width=\figurewidth]{./fig/cifar/erf_unknown_sample_low_uncertainty/image\x.png}};

      \foreach \x [count=\i] in {True: dog\\cat: 1.00,True: dog\\cat: 1.00,True: ship\\plane: 1.00,True: dog\\cat: 1.00,True: dog\\cat: 1.00,True: dog\\cat: 1.00,True: dog\\cat: 1.00}
        \node[align=center]
          (\i) at ({\figurewidth*\i},{-\figureheight*1.75})
          {\tiny \x};          
          
    \end{tikzpicture}}
   {Highest/lowest variance test samples (ERF activation)}    
  \end{subfigure}
  \caption{Results with ERF activation}
  \end{subfigure}
  \\[4em]
  \begin{subfigure}{\textwidth}
  \begin{subfigure}[t]{.49\textwidth}
    \centering
    \input{./fig/cifar/cifar10_step.tex}\\[0em]
  \end{subfigure}
  \hfill
  \begin{subfigure}[t]{.49\textwidth}
    \centering
    \setlength{\figurewidth}{.18\textwidth}
    \setlength{\figureheight}{\figurewidth}
    \resizebox{\textwidth}{!}{
    \begin{tikzpicture}[inner sep=0]

      \tiny

      \node[rotate=90,text width=1.5\figureheight,align=center] at (.25,.35) {\tiny Highest uncertainty (known classes)};
      \node[rotate=90,text width=1.5\figureheight,align=center] at (.25,-1.65) {\tiny Lowest uncertainty (unknown classes)};

      \foreach \x [count=\i] in {True: deer\\plane: 0.30,True: bird\\plane: 0.29,True: cat\\plane: 0.32,True: cat\\plane: 0.26,True: deer\\plane: 0.29,True: deer\\plane: 0.29,True: bird\\plane: 0.32}
        \node[align=center]
          (\i) at ({\figurewidth*\i},{\figureheight*0.75})
          {\tiny \x};

      \foreach \x [count=\i] in {0,1,2,3,4,5,6}
        \node[draw=white,fill=black!20,minimum size=\figurewidth,inner sep=0pt]
          (\i) at ({\figurewidth*\i},{-\figureheight*0})
          {\includegraphics[width=\figurewidth]{./fig/cifar/step_known_sample_high_uncertainty/image\x.png}};

      \foreach \x [count=\i] in {0,1,2,3,4,5,6}
        \node[draw=white,fill=black!20,minimum size=\figurewidth,inner sep=0pt]
          (\i) at ({\figurewidth*\i},{-\figureheight*1})
          {\includegraphics[width=\figurewidth]{./fig/cifar/step_unknown_sample_low_uncertainty/image\x.png}};

      \foreach \x [count=\i] in {True: truck\\car: 0.28,True: ship\\car: 0.25,True: truck\\car: 0.29,True: dog\\plane: 0.31,True: truck\\car: 0.27,True: truck\\car: 0.27,True: frog\\plane: 0.30}
        \node[align=center]
          (\i) at ({\figurewidth*\i},{-\figureheight*1.75})
          {\tiny \x};          
          
    \end{tikzpicture}}
   {Highest/lowest variance test samples (step activation)}    
  \end{subfigure}
  \caption{Results with step activation}
  \end{subfigure}
  \\[4em]
  \begin{subfigure}{\textwidth}
  \begin{subfigure}[t]{.49\textwidth}
    \centering
    \input{./fig/cifar/cifar10_relu.tex}\\[0em]
  \end{subfigure}
  \hfill
  \begin{subfigure}[t]{.49\textwidth}
    \centering
    \setlength{\figurewidth}{.18\textwidth}
    \setlength{\figureheight}{\figurewidth}
    \resizebox{\textwidth}{!}{
    \begin{tikzpicture}[inner sep=0]

      \tiny

      \node[rotate=90,text width=1.5\figureheight,align=center] at (.25,.35) {\tiny Highest uncertainty (known classes)};
      \node[rotate=90,text width=1.5\figureheight,align=center] at (.25,-1.65) {\tiny Lowest uncertainty (unknown classes)};

      \foreach \x [count=\i] in {True: cat\\cat: 1.00,True: bird\\bird: 1.00,True: bird\\bird: 1.00,True: plane\\plane: 1.00,True: deer\\deer: 1.00,True: deer\\deer: 1.00,True: bird\\bird: 1.00}
        \node[align=center]
          (\i) at ({\figurewidth*\i},{\figureheight*0.75})
          {\x};

      \foreach \x [count=\i] in {0,1,2,3,4,5,6}
        \node[draw=white,fill=black!20,minimum size=\figurewidth,inner sep=0pt]
          (\i) at ({\figurewidth*\i},{-\figureheight*0})
          {\includegraphics[width=\figurewidth]{./fig/cifar/relu_known_sample_high_uncertainty/image\x.png}};

      \foreach \x [count=\i] in {0,1,2,3,4,5,6}
        \node[draw=white,fill=black!20,minimum size=\figurewidth,inner sep=0pt]
          (\i) at ({\figurewidth*\i},{-\figureheight*1})
          {\includegraphics[width=\figurewidth]{./fig/cifar/relu_unknown_sample_low_uncertainty/image\x.png}};

      \foreach \x [count=\i] in {True: ship\\plane: 0.76,True: frog\\bird: 1.00,True: truck\\bird: 0.94,True: ship\\plane: 1.00,True: dog\\cat: 1.00,True: frog\\bird: 0.83,True: frog\\cat: 1.00}
        \node[align=center]
          (\i) at ({\figurewidth*\i},{-\figureheight*1.75})
          {\x};          
          
    \end{tikzpicture}}
    {Highest/lowest variance test samples (ReLU activation)}   
  \end{subfigure}
  \caption{Results with ReLU activation}
  \end{subfigure}
  \\[1em]
  \caption{Additional OOD example on CIFAR-10 with 5 classes (`known') in training and all 10 in testing. Left:~Predictive variance histograms for know/unknown test class inputs, where the Mat\'ern shows gradually more separation with decreasing $\nu$, except for the exponential activation which shows poor separation. Right: Test samples from ends of the histograms (true and predicted label + class prob.). For the Mat\'ern-3/2, Mat\'ern-5/2, RBF, and ERF uncertain known and certain unknown samples feel intuitive (good calibration), while the results seem arbitrary for the exponential, ReLU and step activations.}  
  \label{fig:cifar-app-2}
\end{figure}

\end{document}